\documentclass[11pt,reqno]{amsart}
\usepackage[top=1.2in, bottom=1.2in, left=1.2in, right=1.2in]{geometry}
\usepackage{amsfonts, amssymb, amsmath, dsfont, mathtools}
\usepackage{mathrsfs} 
\usepackage[linktoc=page, colorlinks, linkcolor=blue, citecolor=blue]{hyperref}
\usepackage{enumerate,commath}

\usepackage{booktabs}
\usepackage{graphicx}
\usepackage{xcolor}
\usepackage{subcaption} 
\usepackage[font=small]{caption} 
\numberwithin{equation}{section}

\DeclareMathOperator{\sign}{sign}
\DeclareMathOperator{\sgn}{sgn}

\DeclareMathOperator{\AND}{AND}
\DeclareMathOperator{\OR}{OR}

\newcommand{\be}{\begin{equation}}
\newcommand{\ee}{\end{equation}}

\def \R {\mathbb{R}}

\def \AA {\mathcal{A}}

\def \TT {\mathcal{T}}
\def \BB {\mathcal{B}}

\newtheorem{theorem}{Theorem}[section]
\newtheorem{proposition}[theorem]{Proposition}

\newtheorem{lemma}[theorem]{Lemma}

\theoremstyle{remark}
\newtheorem{remark}[theorem]{Remark}

\begin{document}

\title{The Quarks of Attention}
\author{Pierre Baldi \and Roman Vershynin}
\date{\today}

\address{Department of Computer Science, University of California, Irvine}
\email{pfbaldi@uci.edu}

\address{Department of Mathematics, University of California, Irvine}
\email{rvershyn@uci.edu}

\begin{abstract}

Attention plays a fundamental role in both natural and artificial intelligence systems.
In deep learning, attention-based neural architectures, such as  transformer architectures, are widely used to tackle problems in natural language processing and beyond. Here we investigate the  fundamental building blocks of attention and their computational properties. Within the standard model of deep learning, we classify  all possible fundamental building blocks of attention in terms of their source, target, and computational mechanism. We identify and study three most important mechanisms:
additive activation attention, multiplicative output attention (output gating), and multiplicative synaptic attention (synaptic gating). The gating mechanisms correspond to multiplicative extensions of the standard model and are used across all current attention-based deep learning architectures.
We study their functional properties and estimate the capacity of several attentional building blocks in the case of linear and polynomial threshold gates. Surprisingly, additive activation attention plays a central role in the proofs of the lower bounds. Attention mechanisms reduce the depth of certain basic circuits and leverage the power of quadratic activations without incurring their full cost. 
\end{abstract}

\maketitle

\setcounter{tocdepth}{1}
\tableofcontents

{\it ``Everyone knows what attention is... It is the taking possession by the mind in clear and vivid form, of one out of what seem several simultaneously possible objects or trains of thought...''} William James, Principles of Psychology (1890).

\section{Introduction}
\label{sec:intro}

Everyone can focus their attention on an image, a sound, or a thought. But what is attention and how does it really work?
Besides James's definition, other standard definitions of attention include: ``{\it the ability to focus selectively on a selected stimulus, sustaining that focus and shifting it at will}''; or, linking attention to awareness: ``{\it the concentration of awareness on some phenomenon to the exclusion of other stimuli}''. 
All such definitions remain very coarse, based on introspective and phenomenological considerations, and define attention in terms of other functionally obscure terms, such as ``focus''  or ``awareness''. Here, in order to better understand attention at the computational level, we study it within the simplified framework of artificial neural networks and deep learning by first identifying the most fundamental building blocks or quarks, using a physics-inspired terminology, and then rigorously analyzing some of their computational properties. 

The motivation for working with artificial neural networks is two-fold. The first motivation is to avoid getting bogged down by the complexity of biological systems. There is of course a substantial literature on the neurobiology and psychophysics of attention (e.g. \cite{itti2005neurobiology,
arnsten2010neurobiology,posner2011cognitive})
pointing to a variety of different phenomena and attention systems, leading some to conclude: ``{\it The word``attention'' is an inadequate, singular term for a multitude of inter-related
processes. We use a host of adjectives to describe attention—-for example, we say that attention can be divided, oriented, sustained, or focused, and many of these descriptions likely reflect underlying, dissociable neural processes. Complicating matters, attentional resources can be allocated to either external stimuli, or to internal stimuli such as thoughts and memories. Furthermore, we often confuse the regulation of attention (a covert behavior) with the regulation of movement (an overt behavior) when discussing an ``attentional disorder''}'' \cite{arnsten2010neurobiology}. In spite of this complexity and diversity of processes, we believe that at the most fundamental level attention mechanisms are built out of a small number of fundamental
operations, which occur on time scales that are fast compared to the time scales for learning and long-term synaptic modifications. For instance, in order to exclude other stimuli, neuronal machinery must exist that is capable of dynamically suppressing the activity of subsets of neurons, or subsets of connections, or both, associated with the non-attended information.
These fundamental operations may be easier to identify and study using artificial neural networks. In particular, one of our goals here is to produce a systematic nomenclature of all such possible operations, within the standard deep learning formalism. While this is not the place to discuss the relationship between artificial and biological neural networks, there is a body of evidence showing that, atleast at  some level, the former can provide useful information about the latter (e.g. \cite{zipser1988back,olshausen1996emergence,
yamins2016using}).

The second motivation, equally or even more important, is that attention plays an increasingly important role in deep learning systems.

In deep learning networks, various attention mechanisms
 such as content-based attention 
\cite{graves2014neural},
speech recognition attention
\cite{chorowski2015attention},
or 
dot product attention 
\cite{luong2015effective},
have been introduced and successfully deployed 
in applications. Many of these  mechanisms were initially developed 
for speech and natural language applications (NLP) (e.g. \cite{attention_bahdanau,bert,gpt2}),
but they are now being adapted to other problems (e.g. \cite{set_transformer,
fenton2020permutationless}). The intuitive idea in NLP applications is that when, for instance, 
translating a sentence from one language to another, the underlying neural algorithm should be able to dynamically shift its focus on the relevant words and context, while filtering out the less relevant ones.
For instance when translating ``the red roof'' into the French ``le toit rouge'' the machinery that produces the {\it third} word of the output (``rouge'') should dynamically give more importance to the {\it second} word (``red'') of the input, relative to the other neighboring words.  
The current pinnacle of attention-based architectures is the transformer architecture
\cite{vaswani2017attention,transformers}
which has led to state-of-the-art performance in NLP and is now widely used.
These advances have even led some experts to speculate that attention mechanisms may be key for achieving machine consciousness (!). 

However, with rare exceptions
\cite{dong2021attention}, there is little theory to help us better understand the nature and computational capabilities of attention.
To address this gap, in Section 2 we first seek to identify and classify the most fundamental building block of all attention mechanisms within the deep learning framework. In particular, we identify three key attentional mechanisms we call activation attention, output gating, and synaptic gating. In Section 3, we show how output gating and synaptic gating are used in all the current attention-based architectures, including transformers. In Section 4, we explore the functional capacity of output gating and synaptic gating. In Section 5, we provide a brief overview of the notion of capacity and the technique of multiplexing, which is a form of activation attention, for proving capacity lower bounds. In Sections 6 and 7, we prove several theorems about the capacity of 
activation, output, and synaptic gating, using multiplexing, first for single units and then for single layers
of linear and polynomial threshold functions.

\section{Sytematic Identification of Attention Quarks: Within and Beyond the Standard Model}
\label{sec:gating}

We first introduce the formal neural network framework that we use in order to systematically organize and study the attention quarks, i.e. the most fundamental building blocks of attention. To borrow another term from physics, we call this framework the Standard Model.

\subsection{The Standard Model (SM)}

The Standard Model is the class of all neural networks made of what are generally called McCulloch and Pitt neurons. Neural networks in the SM consist of directed weighted graphs of interconnected processing units, or neurons. The synaptic strength of the connection from neuron $j$ to neuron $i$ is represented by a single real-valued number $w_{ij}$. Any non-input neuron $i$ produces an output $O_i$ by first computing an activation $S_i=\sum_j w_{ij}O_j$, i.e the activation corresponds to the dot product of the incoming signal with the synaptic weights. In turn, the output of the neuron is produced in the form $O_i=f_i(S_i)$ where $f_i$ is the transfer or activation function of neuron $i$. Typical activation functions include the identity function in the case of linear neurons, sigmoidal activation functions such as the logistic and tanh activation functions, and piece-wise linear functions (\cite{tavakoli2021splash}), such as the Heaviside, sign, or ReLU functions. An encompassing, and more than sufficient, class of transfer functions for a formal definition of the SM is the class of functions that are differentiable everywhere except for a finite (and small) set of points. A fundamental, and easy to prove \cite{baldi2021deep}, property of the SM is that it has universal approximation properties: (1) any Boolean function can be implemented exactly by a feedforward network in the SM; and (2) for any small $\epsilon >0$, any continuous function from 
$\R^n$ to $\R^m$ defined on a compact set $C$ can be 
approximated within $\epsilon$ everywhere over $C$ by a feedforward network in the SM.

Several attention mechanisms described below can be viewed as extensions of the standard model, where new operations are added to the SM to obtain a richer model. Extending the SM is not a new procedure. For instance, using softmax layers is already an extension of the SM since 
the softmax is not a proper, single-neuron, activation function. Another example is the use of polynomial activation functions 
(e.g. \cite{baldi2019polynomial}).
Due to the universal approximation properties of the SM, these extensions are not meant to increase the approximating power of the SM. 
Rather, their value must be established along other dimensions, such as circuit size or  learning efficiency.
In the digital simulations of neural networks, these extensions correspond to new software primitives. In physical neural networks, these extensions must come with actual wires and physical mechanisms. For instance, a softmax operation is a new software primitive in a neural network software library but it requires a new physical mechanism for its physical implementation. It can be replaced by a network of depth 3 within the SM (Section \ref{sec:functional} with fixed weights set to $\pm 1$ Figure \ref{fig:SoftMax}), provided logarithm and exponential activation functions are available.

\subsection{Systematic Taxonomy}

In the SM, there are three kinds of variable types: $S$ (activations), $O$ (outputs), and $w$ (synaptic weights).
At the most fundamental level, we can organize attention mechanisms (and more broadly new SM interactions) depending on:
the type of variable associated with the {\it source} of an attention signal (3 possibilities), the type of variable associated with the {\it target} of an attention signal (3 possibilities), and on the {\it mechanism} of the interaction, i.e. on the algebraic operation used to combine the attending signal and the attended target. While many algebraic operations can be considered, the two most basic ones are addition and multiplication (two possibilities)--resulting in a total of 18 different possibilities. These could be further subdivided depending on multiplicity issues, at both the source and the target, as well as time scales, as described below. We now discuss these possibilities, reducing them down to the 6 most important ones. 

\begin{enumerate}
\item{\bf Source:} It is reasonable to assume that the source of the attending signal is a variable of type $O$ corresponding to the output of one attending neuron, or a group (layer) of attending neurons. While other possibilities can be explored, e.g. a synapse directly attending another synapse,
they would require new complex mechanisms in a physical implementation. Furthermore, they do not occur in current attention-based deep learning models. The same can be said for the activation being the direct source of the attending signal. Even more unlikely would be the case of mixed schemes where the attending signal would emanate, for instance, from both neuronal outputs and synapses. In short, the reasonable assumption that the attending signals emanate from neuronal outputs allows us to reduce the number of possibilities by a factor of three leaving 6 basic possibilities (Table \ref{tab:taxonomy}.

\item {\bf Target:} For the target of an attention signal, we will study all three possibilities. Thus attention signals can target activations ($S$), outputs ($O$), or synapses ($w$). We will call these three forms of attention activation attention, output attention, and synaptic attention respectively.

\item {\bf Mechanism:} The most simple operations one can think of for combining the attending signal with its attended target are addition and multiplication. 
Attention requires 
excluding all other stimuli and possibly enhancing  the attended stimulus (here we do not distinguish between external stimulus or internal representation). Intuitively, at the fundamental level, these exclusions and enhancements correspond to multiplicative operations where, for instance, the signals associated with non-attended stimuli are inhibited--i.e. multiplied by zero, and the attended stimuli are enhanced, i.e. mutliplied by a factor greater than one. 
We will reserve the term ``gating'' for multiplicative interactions. 
Thus, for instance, multiplicative synaptic attention will also be called synaptic gating.
All multiplicative interactions, with the exceptions of terms of the form $w_{ij}O_j$, are not part of the SM and thus correspond to potential extensions of the SM.

However, for completeness, we will also consider the case of additive interactions. Furthermore, in the case of activation attention, for several common activation functions such as logistic or ReLU, inhibition (and thus suppression of stimuli) can be achieved additively by sending a large negative signal towards the 
attended neuron. Unlike gating, additive activation attention is contained in the SM.  
Note that both addition and multiplication are differentiable operations, and thus can easily be incorporated into the backpropagation learning framework.

\item {\bf Multiplicities:}
In each possible case, one must take into account multiplicity issues both at the level of the source and at the level of the target. For instance, in synaptic gating, can the attending output of a neuron gate more than one synapse? Can the attending output of several neurons gate the same synapse? And so forth. In the most simple cases, we will assume that the multiplicity is one both at the source and at the target, but greater multiplicities will also be considered, for instance in some of the theorems in Sections \ref{sec:capacitysingle} and \ref{sec:capacitylayers}. 

\item {\bf Time Scales:} Finally, for simplicity, and in line with current deep learning attention models, we assume that the attention mechanisms operate on the time scale of individual inputs. Different inputs create different attention signals.
Alternative possibilities are briefly discussed in Section 
\ref{sec:SVO}).
\end{enumerate}

In summary, we are left with six main cases, corresponding to two different mechanisms $(+,\times)$ and three different target types $(S,O,w)$. We now examine them one by one and show that they can be reduced to three most important cases, which are further studied in the following sections. 
Finally, for each case, it is useful to keep in mind the difference between digital simulations and actual implementations in a physical neural network, i.e. machine learning versus learning in the machine \cite{baldi2021deep}. 
For instance, different mechanisms may be equivalent at the level of the algebraic expressions they lead to, but very different in terms of their physical implementations.

\subsection{Identification: Additive Interactions}

In the case of additive interactions, the attention signal is added to three possible targets of type $S$, $O$, or $w$.

\subsubsection{Additive Activation Attention: Multiplexing}

In this case, consider an attended neuron $i$. It activation
$S$ has the form $S=S_1+S_2$ where $S_1$ is the ``normal'' activation (without attention) and $S_2$ is the attending signal originated from one, or multiple, attending neurons. The terms multiplexing simply refers to the combination or superposition of two signals over the same channel.
Depending on the transfer function $f_i$ of neuron $i$, the attending signal can be used to control the output $O_i=f_i(S_1+S_2)$. The typical case is when $f_i$ is the logistic or Heaviside function: then a large negative signal $S_2$ (much larger than $S_1$) will override any $S_1$ and
force the output of neuron $i$ to be zero. If, on the other hand, $S_2=0$ then $O_i=f_i(S_1)$ and the normal signal will be propagated. If attention must be able to both suppress and enhance signals, this mechanism allows the suppression, but it does not provide a direct way for the multiplicative enhancement of signals. Formally this mechanism is entirely within the SM and does not require extending it. If the attending signal must come from a single neuron (source multiplicity one), this can easily be achieved by connecting the output of the attending neurons to a single linear neuron
whose output is equal to $S_2$. Although not new, this attention mechanism is interesting because it will play a central role in the methods for proving various technical results about the new gating mechanisms presented below. 

\subsubsection{Additive Output Attention}

In this case, using multiplicities of 1, we
consider a neuron $i$ connected to a neuron $k$ in the main network, and an attending neuron $j$. In this case, the output $O_j$ is simply added to the output $O_i$
(Figure \ref{fig:addition}, Left) producing the terms
$O_i+O_j$ (or $O_i+w_{ij}O_j$). This terms is nothing new in the SM and is equivalent to having an additional linear neuron with two incoming connections originating in neurons $i$ and $j$, both with synaptic weight 1, and the same outgoing connections as neuron $j$ in the original network. This mechanism alone does not provide much in terms of  attentional functionalities and therefore it will not be considered here any further.

\subsubsection{Additive Synaptic Attention}

\begin{figure}[ht]
\begin{center}
\includegraphics[width=1.0\columnwidth]{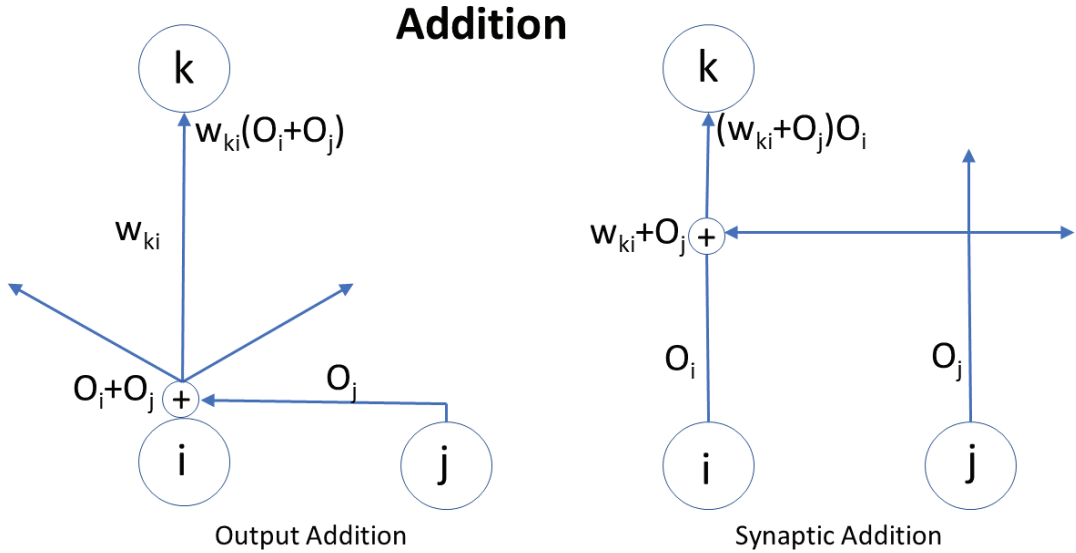}
\end{center}
\vspace{-0.2cm}
\caption{Additive Output (Left) or Synaptic (Right)  Interactions. Left: the output $O_j$ of the attending neuron is added to the output $O_i$ of the attended neuron, producing a term of the form $O_i+O_j$. Right:
the output $O_j$ of the attending neuron is added to the attended synaptic weight $w_{ki}$, {\it de facto} producing a new multiplicative term of the form $O_iO_j$ as one of the input components to neuron $k$.}
\label{fig:addition}
\end{figure}

In this case, using multiplicities of 1, we
consider a neuron $i$ connected to a neuron $k$ in the main network, and an attending neuron $j$. In this case, the output $O_j$ is simply added to a synaptic weight, i.e. to $w_{ki}$.
(Figure \ref{fig:addition}, Right), producing a new 
synaptic weight 
$w_{ki} +O_j$, which in turn creates a contribution equal to $(w_{ki}+O_j)O_i$ st neuron $k$. This contribution contains a new multiplicative term of the form $O_iO_j$ which is not part of the SM.
Since $O_iO_j$ falls under the multiplicative category, it is subsumed by the analyses below of multiplicative interactions; thus additive synaptic interactions will not be considered any further in the rest of this work.
 
In summary, there are three kinds of additive interactions.
Only multiplexing (additive activation attention) will be used in the rest of this work, and primarily as a tool in the proofs of some theorems.

\subsection{Identification: Multiplicative Interactions or Gating}

In the case of multiplicative interactions, the attention signal is multiplied with three possible targets of type $S$, $O$, or $w$.

\subsubsection{Multiplication Activation Attention: Activation Gating}

In this case, using a source multiplicity of 1, consider 
an attended neuron $i$ with activation $S_i=\sum w_{il}O_l$ and transfer function $f_i$ and an attending neuron $j$. In this case, the attending signal $O_j$ multiplies the activation $S_i$ so that the final output of neuron $i$ becomes $O_i=f_i(S_iO_j)$. If $f_i$ is sigmoidal or a threshold function, a large positive value of the attention signal $O_j$ could be used to drive the response of neuron $i$ towards one of its extreme values (e.g. $0/1$ or $-/+$ \footnote{Everywhere we write -/+ to indicate 
-1/+1.} ).
Note that $O_i=f_i(S_iO_j)=f_i(\sum_l w_{il}O_lO_j$, so formally this mechanism is equivalent to having $O_j$ multiply the output $O_l$ of all the neurons connected to neuron $i$, although in a physical implementation these two things could be very different. Because of this equivalence, we will consider that output gating subsumes this mechanism and we will not discuss it much further. Furthermore, at least in the case of attended and attending neurons with threshold transfer functions equal to the sign function, multiplication of activation and multiplication of output are directly equivalent at the algebraic level because:
$\sign (S_i O_j)=\sign(S_i) \sign(O_j) =\sign S_i O_j=O_iO_j$.

\subsubsection{Multiplicative Output Attention: Output Gating}

\begin{figure}[ht]
\begin{center}
\includegraphics[width=0.9\columnwidth]{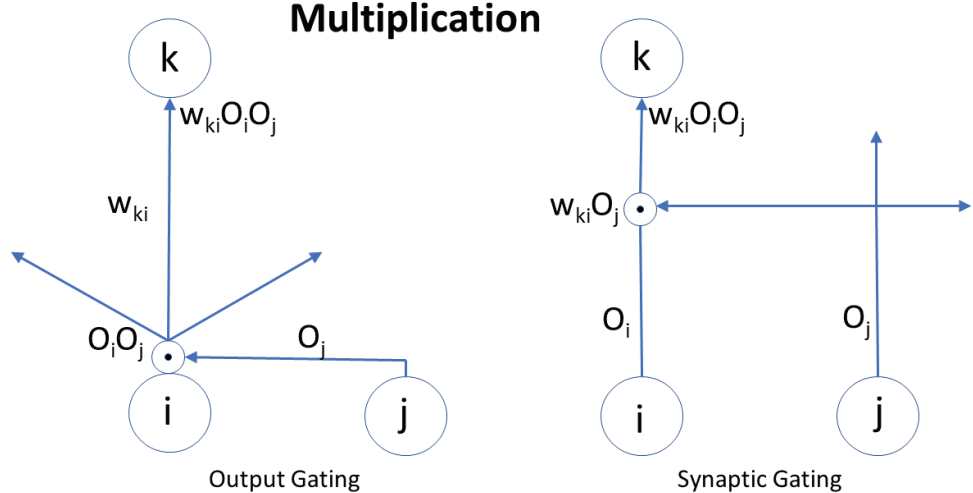}
\end{center}
\vspace{-0.2cm}
\caption{Multiplicative Interactions: Output and Synaptic Gating. Left: In output gating, neuron $j$ gates the output of neuron $i$ producing a new effective output $O_iO_j$. The signal $O_iO_j$ is  broadcasted to all the neurons downstream of neuron $i$, including neuron $k$. Right: In synaptic gating, neuron $j$ gates the synapse between neuron $i$ and neuron $k$, producing a new effective synaptic weight equal to $w_{ki}O_j$. In both cases, the signal $O_j$ can be transmitted to other neurons and other synapses (higher multiplicity). In both cases, neuron $k$ receives the same signal $w_{ki}O_iO_j$. However the effects of output versus synaptic gating on the rest of the network are different (see text).}
\label{fig:gating}
\end{figure}

In this case, using multiplicities of 1, we
consider a neuron $i$ connected to a neuron $k$ in the main network, and an attending neuron $j$. In this case, the output $O_i$ is multiplied by $O_j$ (or $w_{ij}O_j$) producing the quadratic terms 
$O_iO_j$ which is new in the SM, leading to an input component into neuron $k$ equal to $w_{ki}O_iO_j$
(Figure \ref{fig:gating}, Left). Note that while the multiplication is commutative, the attention mechanism is not in the sense that only the axon emanating from neuron $i$ carries the signal $O_iO_j$ to all the targets of neuron $i$.

\subsubsection{Multiplicative Synaptic Attention: Synpatic Gating}

In this case, using multiplicities of 1, we
consider a neuron $i$ connected to a neuron $k$ in the main network, and an attending neuron $j$. In this case, 
the synaptic weight $w_{ki}$ is multiplied by $O_j$.
This produces a new synaptic weight 
$w_{ki}O_j$, which in turn also  creates a contribution equal to $w_{ki}O_iO_j$ into neuron $k$.

\subsection{Synaptic Gating versus Output Gating}
\label{sec:SVO}
\begin{figure}[ht]
\begin{center}
\includegraphics[width=0.9\columnwidth]{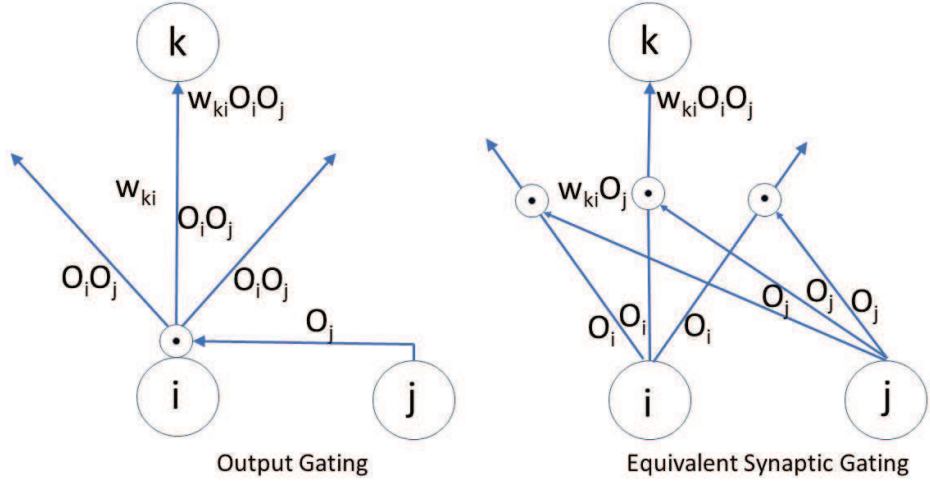}
\end{center}
\vspace{-0.2cm}
\caption{Output gating equivalence. Left: Output gating of neuron $i$ by neuron $j$. All the connections emanating from neuron $i$ carry the signal $O_iO_j$. Right: Equivalent network obtained using synaptic gating only. The gating neuron $j$ must synaptically gate all the connection weights emanating from neuron $i$}.
\label{fig:gating2}
\end{figure}

\begin{figure}[ht]
\begin{center}
\includegraphics[width=0.9\columnwidth]{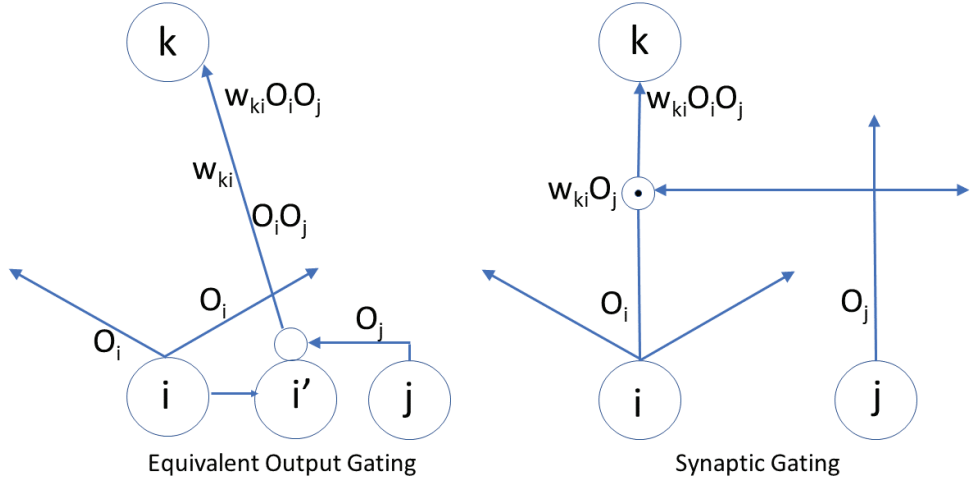}
\end{center}
\vspace{-0.2cm}
\caption{Synaptic Gating Equivalence. Right: Synaptic gating of synaptic weight $w_{ki}$ by neuron $j$.
Left: Equivalent network obtained using output gating only. Neuron $i$ has an identical twin neuron $i'$, i.e. neuron $i$ connected to neuron $i'$ through the identity function and thus both neurons produce the same output equal to $O_i$.
Neuron $j$ output gates neuron $i'$ producing a signal $O_iO_j$ which  travels through the synapse $w_{ki}$.
All other connections emanating from neuron $i$ carry the signal $O_i$ and are unaffected by the gating neuron $j$.
}
\label{fig:gating3}
\end{figure}

When the gating signal $O_j$ is close to zero, it will tend to suppress the gated signal $O_i$ or the gated synaptic weight  $w_{ki}$.
The ability to dynamically suppress a synaptic weight or the signal flowing through it embodies the idea of 
``excluding other stimuli'' associated with attention. Likewise, when the gating signal $O_j$ is far from zero, it can  dynamically enhance a synaptic weight or the signal flowing through it. 
Although equivalent circuits for output and synaptic gating can be found
(see Figures \ref{fig:gating2} and \ref{fig:gating3}), conceptually they are different.

Synaptic gating is a mechanism by which the gating  neuron or network can dynamically change the synaptic weights of the gated neuron or network, thus effectively changing the program being executed by the attended network. 
This allows the same gated neuron or network to be modulated and to compute different functions, as a function of the gating neuron or network.
Thus synaptic gating can also be viewed as a form of fast synaptic weight mechanism 
\cite{schmidhuber1992learning,ba2016using}, where synapses with different time scales coexist and fast synapses are used to dynamically store information and modulate the function being computed by a given network. However, even for the fast synapses there could be different time scales. While here we assume that synapses change on the time scales of the inputs, fast synapses could also change on a lower time scale in the sense that a gated synapse could be reused over several inputs. 

Although we have seen that both synaptic and output gating produce the same term of
the form $w_{ki}O_iO_j$ at neuron $k$, this is true only for neuron $k$.
Unlike synaptic gating, 
output gating affects all the neurons downstream of the gated neuron. In contrast, synaptic gating is more precise as it affects only the neuron downstream of the gated synapse, but it is more expensive, requiring one gating wire per gated synapse, rather than one gating wire per gated neuron.
Nevertheless, if neuron $i$ has only one outgoing connection, then gating of its output or its outgoing synapse are of course equivalent. 
For this reason, in the formal analyses, we will focus on output gating which covers also synaptic gating under the assumption of a single outgoing connection per gated neuron.

An observation that will become important in Sections \ref{sec:functional}-- \ref{sec:capacitylayers}, is that in the case of binary units and output gating, it does make a difference whether one uses $0/1$ or $-/+$ representations. In particular, although $0/1$ or $-/+$ linear (or polynomial) threshold functions are equivalent, different forms of output gating are obtained with different combinations of such units. This is because multiplication of $x \in \{-1,0,1\}$
by $0$ or by $-1$ leads to different results. In particular, multiplication of the outputs of two $0/1$ threshold gates is equivalent to applying a logical AND operation, whereas multiplication of two $-/+$ linear threshold gate is equivalent to applying a logical NXOR (the negation of an XOR). Multiplication of a $0/1$ threshold gate by a $-/+$  threshold gate produces a non-Boolean functions with outputs in $\{-1,0,1\}$. 
Nevertheless in many cases equivalent circuits can be found (see Example in Figure \ref{fig:XOR}) using either multiplication between $0/1$ threshold gates or multiplication between $-/+$ threshold gates. This also suggests a more general question of studying all possible ways of combining two threshold functions using Boolean operators (see Section
\ref{sec:capacitysingle}).

\subsection{Relations to Polynomial Neural Networks}
There are at least two important relationships between gating and polynomial neural networks. First, we have seen that both synaptic and output gating mechanisms produce quadratic terms of the form $w_{ki}O_iO_j$ contributing to the activation of neuron $k$.
Thus gating can also be viewed as a special case of neurons with quadratic activations or, more generally, polynomial activations \cite{baldi2019polynomial}.
However, a full quadratic activation function of $n$ inputs may need $n(n-1)/2$ 3-way synaptic weights (the quadratic component of the activation of a neuron $i$ has the form 
$S_i=\sum_{jk} w_{ijk} O_jO_k$)
associated with each possible pair of inputs. 
Synaptic gating or output gating produce only one new quadratic term. 
Thus, in short, gating creates quadratic terms but in a sparse way that avoids the combinatorial explosion associated with all possible combinations.

The second connection is that the same gating concepts can be applied to
to more complex units, beyond the standard model, in particular to units where the activation is a polynomial function of degree $d$ of the inputs (the standard model corresponds to $d=1$).
Thus for instance a neuron $j$ with a quadratic activation function could gate the output of another neuron $i$ with quadratic activation functions, or gate a synapse $w_{ki}$ between neuron $i$ and neuron $k$. Gating by neurons with polynomial activations, in particular gating by polynomial threshold units, will be studied in Sections \ref{sec:capacitysingle} and \ref{sec:capacitylayers}.

\begin{table}
  \caption{Organization of attention mechanisms. Assuming that the 
  origin of the attention signal is the output of one or several neurons, there are 6 classes depending on the target of the signal and the interaction mechanism. We consider 3 kinds of targets: activation ($S$), output ($O$), and synapses ($w$). We consider 2 kinds of interaction mechanisms: addition and multiplication. 
Two of the classes (additive activation attention, or multiplexing, and additive output attention) are in the SM; the other 4 classes correspond to true extensions of the SM. The discussion in the text shows that further analyses can focus on three classes only: multiplexing, output gating, and synaptic gating (in bold). 
 }
  \label{tab:taxonomy}
  \centering
  \begin{tabular}{|l|c|c| c|}
    \hline
     &$S$ &$O$& $w$   \\ \hline 
Addition & {\bf multiplexing} (SM) & additive output  att.(SM)&  aditive synaptic att. \\ \hline
Multiplication & activation gating & {\bf output gating} & {\bf synaptic gating}\\ \hline
  \end{tabular}
\end{table}

\subsection{Summary}

In summary, the quarks of attention can be classified based on the origin, the target, and the interaction mechanism of the attention signal. Assuming that the origin is in the output of one neuron, or a group of neurons, and that the interactions are either additive or multiplicative, this leads to six classes (Table \ref{tab:taxonomy}). Within the additive group, two classes are already in the SM (additive activation and additive output attention) and only one class is of interest here for further studies (additive activation attention or multiplexing). Within the multiplicative group, all three classes correspond to true extensions of the SM and, at least formally, further analyses can be reduced to two main classes: output gating and synaptic gating. In all cases, the attending signal modulates the function 
computed by the attended network.


\section{All you Need is Gating:  Transformers}
\label{sec:transformers}

\begin{figure}[ht]
\begin{center}
\includegraphics[width=1.2\columnwidth]{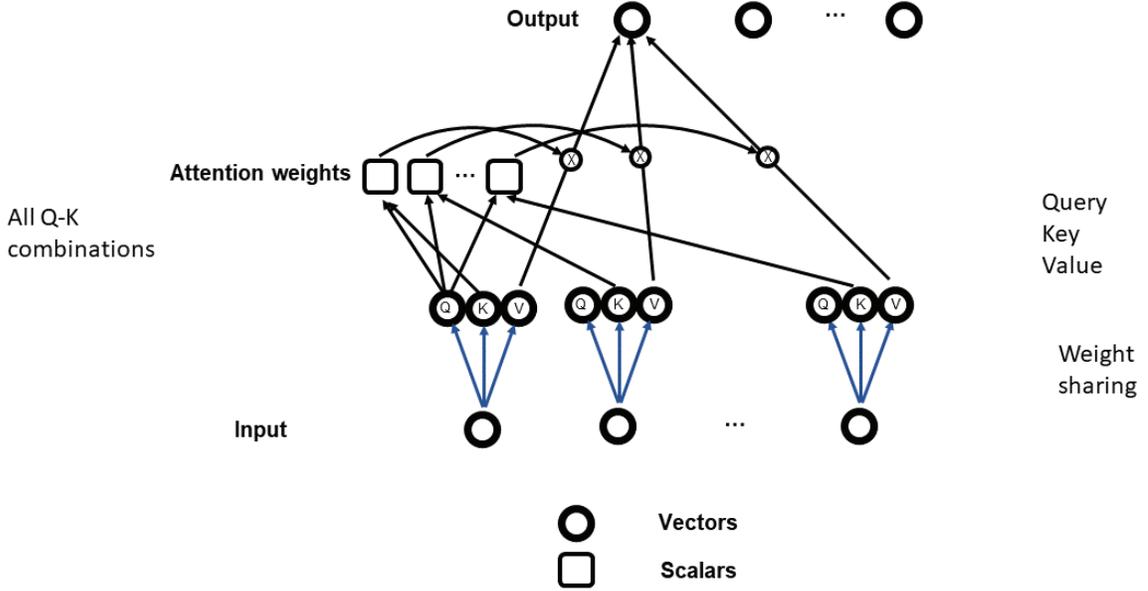}
\end{center}
\vspace{-0.3cm}
\caption{Neural network representation of the basic encoder module of a transformer architecture. Each input vector is converted into three vectors Q (Query), K (key), and V (value) using weight sharing (blue weights). All $n^2$ pairwise dot products $K(k)Q(l)$ are computed in the attention layer, which corresponds to a set of output gating operations. This is followed by row-wise softmax operations on these dot products to produce the weights that are used to linearly combine the value vectors  into each corresponding output. These linear combinations correspond to synaptic gating. }
\label{fig:transformer}
\end{figure}

Although the descriptions of attention mechanisms in deep learning often seem complex and sometimes obscure the
underlying neural architecture
\cite{graves2014neural,
chorowski2015attention,
luong2015effective,
attention_bahdanau,bert,gpt2},
it can be checked that in all cases these are built out of the output and synaptic gating operations described in the previous section. For conciseness, here we  demonstrate this in detail only for the transformer architectures
\cite{vaswani2017attention,transformers}
(see also \cite{liu2021pay} for an MLP alternative to transformers). 
These architectures consist of stacks of similar encoder and decoder modules, with attention mechanisms in each module.
The details of an encoder module are shown in Figure 
\ref{fig:transformer}. As the Figure shows, a shared and typically linear network is first applied to each of $n$ input vectors. At the bottom of the architecture, these input vectors
could represent for instance vectors encoding successive words from a sentence. At higher levels of the stack, these vectors could be associated with the outputs of the previous encoder or decoder module and correspond to more abstract representations. 
For each input vector, the shared network typically produces a triplet of vectors of the same size $m$: $Q$ (Query),$ K$ (key), and $V$ (value), for a total of $3n$ triplets. The subsequent attention mechanism is drawn in a concise way in the Figure and is based on three operations: (1) taking all $n^2$ pairwise dot products of the $n$ query vectors with the $n$ key vectors; (2) applying a softmax to each row of dot products; and (3) using the output of the softmax operations as weights for linearly combining the value vectors to produce the corresponding output vector at each position. 
The first operation can be built using output gating, each dot product involving $m$ gating operations, to multiply the proper $Q$ and $K$ components together. As a side note, these dot products can be viewed as similarity measures between the $Q$ and $K$ vectors, especially when these are normalized, and this suggests other kinds of transformer architectures where different similarity kernels are used. 
The softmax operation is a standard extension of the SM (Figure \ref{fig:SoftMax}). The third operation corresponds to synaptic gating of the connections between the $V$ vectors and the outputs. The convex combination of the value vectors by the corresponding softmax weights determines how much each value vector influences each output vector, based on the corresponding similarities between $Q$ vectors and $K$ vectors. This is where the influence of some of the value vectors can be enhanced, while the influence of others can be suppressed.
Thus in total there are $mn^2$ output gating operations, and 
$n^2$ synaptic gating operations (assuming $n$ output vectors).
Thus, in short, the entire encoder module is based on a large number ($O(mn^2)$) of gating operations, both of the output and synaptic type. Thus, in this form, it can only be applied when $n$ is not very large. The basic transformer decoder module (not shown) is very similar. One important property of the encoder module conferred by the attention mechanisms is that the output is invariant under permutation of the inputs. This is because any permutation of the inputs, results in a corresponding permutation of the Q,K, and V vectors due to the weight sharing. This in turn induces a corresponding permutations in the dot products and softmax outputs, so that in the end the weighted contribution of any V vector into any output vector remains the same. This may seem surprising for an architecture that was originally developed for NLP tasks, where the order of the words obviously matter. Indeed, very often in practice positional information is added to each input vector. The permutation invariance of transformers is particularly beneficial for applications of transformers outside of NLP, in particular applications where the input consists of {\it sets} of data vectors, where the order of the data vectors does not matter (e.g. \cite{set_transformer,
fenton2020permutationless}).

\section{Functional Aspects of Attention}
\label{sec:functional}

Next we study though several examples how certain functionalities can be implemented using attention mechanisms, beginning with the effect of attention on single units.

\subsection{Single Unit Output Gating: Shaping the Activation Function}

\begin{figure}[ht]
\begin{center}
\includegraphics[width=0.75\columnwidth]{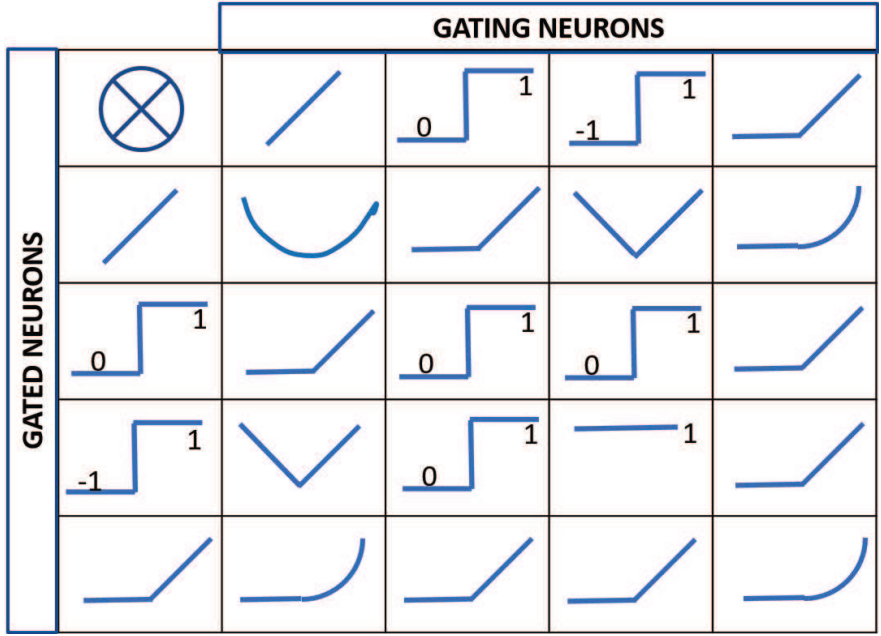}
\end{center}
\vspace{-0.1cm}
\caption{Effect of gating on activation functions. For simplicity we consider four main activation functions: linear, threshold (0,1) [Heaviside function], threshold (-1,1) [sign function], and ReLU.}
\label{fig:activation}
\end{figure}

First, for simplicity, we consider output gating of a unit by another unit with the same inputs and the same weights, hence the same activation $S$. The two units may have two different activation functions $f$ and $g$. Through output gating, the final output of the gated unit will be given by:
$f(S)g(S)=fg(S)$. Thus, in this case, output gating is equivalent to changing
the activation function of the gated unit
from $f$ to $fg$. Examples of this effect
are shown in Figure (Figure 
\ref{fig:activation}) where both $f$ and $g$ are piecewise linear, and centered at 
the origin. Note that in the case of a linear unit gated by another linear unit, the final output is a quadratic function of the $n$ inputs, but with only $O(n)$ parameters as opposed to $O(n^2)$. The ReLU activation function emerges naturally, through the gating of a linear function by a $(0,1)$ threshold function, or vice versa. Finally, the symmetric wedge activation function \cite{tavakoli2021splash} emerges also naturally through the gating of a linear function by a $(-1,1)$ threshold function.

\subsection{Single Unit Attention: XOR}

\begin{figure}[ht]
\begin{center}
\includegraphics[trim={0 0 0 1.2cm},width=1.0\columnwidth]{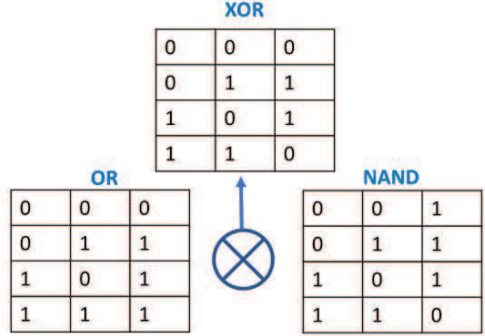}
\end{center}
\vspace{-1.9cm}
\caption{Shallow computation of XOR by a single unit with attention. The left unit computes an OR, which can be implemented by a single linear threshold gate. The right unit computes a NAND, which can also be implemented by a single linear threshold gate. The gating of one unit by the other produces the XOR function. The XOR function cannot be implemented using a shallow (no hidden layer) network of 
 linear threshold gates. In this particular examples, the gated and the gating $0/1$ units can easily be replaced by $-/+$ units, since it is always possible to linearly separate any point on a hypercube from all the other points with a hyperplane, and the component wise product of $(-1,1,1,1) \times (1,1,1,-1)$ gives
 $(-1,1,1,-1)$, which corresponds to the $-/+$ version of XOR.}
\label{fig:XOR}
\end{figure}

Next, we look at the simple XOR function.
It is easy to show that the XOR function cannot be computed in a shallow way by a single linear threshold gate (or sigmoidal) neuron. Its computation requires at least one hidden layer.
However, as shown in Figure
\ref{fig:XOR} using 0/1 outputs, the XOR function can be computed by a shallow network with a single linear threshold unit output-gated by another linear threshold unit. To 
see this, note that any corner of the hypercube can always be isolated by a hyperplane from the other corners of the hypercube, i.e. there is always a linear threshold gate that has value 1 (resp. 0) for one Boolean setting of its inputs, and 0 (resp. 1) for all the other possible inputs (see Lemma \ref{lm:multi}). In particular, both the OR and NAND
functions are of this kind and thus can be implemented by a linear threshold gate. Gating the OR by the NAND (or vice versa) produces the desired XOR function without using any hidden layers, assuming that the ouput gating operation is an integral part of the layer where it occurs.

\subsection{Attention Layers: Universal Approximation Properties} 

Next, we look at how universal approximation proofs are affected if output gating is allowed, both in the Boolean and continuous cases. 

\subsubsection{The Boolean Case.}

Every Boolean function of $n$ variables can be computed by a feedforward network of linear threshold gates, since AND, OR, and NOT can be implemented by linear threshold gates. By expressing the function in disjunctive or conjunctive normal form, the implementation can be achieved with a single hidden layer of exponential size. If we allow output gating, and its iterations, we have the following theorem. 

 \begin{proposition}
 \label{prop:universal}
 Every Boolean function of $n$ variables can be expressed as the product of at most $2^{n-2}$ linear threshold gates, both in the $0/1$ and $-/+$ representations. 
 \end{proposition}
 
 \begin{proof}
 Let $f$ be a Boolean function of $n$ variables, using $0/1$ to denote false and true respectively. If $f$ is 0 everywhere, it can immediately be expressed as a linear threshold gate. Likewise, if $f$ is 0 everywhere but one point of the $n$ dimensional cube, then it can be immediately expressed as a single linear threshold gate. Thus we can assume that $f$ is 0 on at most $2^{n-2}$ points. Let 
 $x_1,\ldots,x_L$($L \leq 2^{n-2}$) denote the inputs where $f$ is zero. For each index $i$, let $g_i$ denote the linear threshold gate which has value $0$ on $x_i$ and $1$ everywhere else. Then it is easy to check that $f(x)$ can be written as the product $f(x)=f_1(x)\ldots   f_L(x)$ (alternatively, on can express $f$ in conjunctive normal form).  The proof is the same in the $-/+$ case, letting $g_i(x)$ be the linear threshold gate with value $-1$ for $x_i$, and $+1$ everywhere else. Obviously the same result holds for polynomial threshold gates of degree $d$.
\end{proof}

In the $-/+$ case, the set $B_n$ of all Boolean functions with the  multiplication operation forms a commutative group, and each Boolean function is its own inverse. 
The subset of all linear threshold gates contains the identity, and each linear threshold function gate is its own inverse. 
However it does not form a subgroup because it is not closed. By the theorem above, the multiplicative closure of the set of all linear threshold gates is the set $B_n$ of all Boolean functions.  

Since every Boolean functions can be written as a product of an exponential number of linear (or polynomial) threshold gates, it is natural to ask whether a smaller number of factors may be used. Can every Boolean function be written as the product of a linear or polynomial number of linear threshold gates?  We will answer this question negatively in Section \ref{seq:negative}.

\subsubsection{The Continuous Case.}
Next we look at the continuous case, using output gating to modify the basic universal approximation proof
\cite{baldi2021deep}. 

\begin{theorem}
 Let $f$ be a continuous function from $[0,1]$ to $\R$, and $\epsilon >0$. Then there exists an integer $n=n(\epsilon)$ such that $f$ can be approximated within $\epsilon$ everywhere over $[0,1]$ by a network of $n$ linear units, attended by $n$ corresponding linear threshold gates with output gating. The final approximation corresponds to the dot product between the vector of linear unit outputs and the vector of attending unit outputs.   
\end{theorem}

\begin{proof}
Since $f$ is continuous over the closed interval, it is uniformly continuous so that there exists $\delta >0$ such that for any $x_1$ and $x_2$ in $[0,1]$:
\be
\vert x_2-x_1\vert < \delta 
\Rightarrow
\vert f(x_2)-f(x_1) \vert < \epsilon
\label{eq:}
\ee
Let us choose an integer $n$ large enough so that $\delta> 1/n$. Next we slice the interval $[0,1] $ into $n$ slices of width 
$1/n$. Next we construct a network with $n$ linear units and $n$ linear threshold gate attention units with outputs in $\{0,1\}$ (the proof can be adjusted to accommodate outputs in $\{-1,1\}$).
All the attention units are connected to the single input $x$ by a weight equal to 1. Their threshold (bias) however are $0, 1/n,2/n, \ldots ,(n-1)/n$ so that when $x \in [0,1/n)$ only the first attention unit is on, when $x \in [1/n,2/n)$ only the first two attention units are on, and so forth. In other words, the slice containing $x$ is encoded in the number of linear threshold gates that are on. 
For the linear units, they compute values $y_1(x), \ldots,y_n(x)$ as follows. The first linear unit approximates the function $f$ in the first slice by producing the line that goes through $f(0)$ and $f(1/n)$, i.e. by implementing the function $y_1(x) =
f(0)+ n[f(1/n)-f(0)]x$. The second linear unit approximates the function $f$ in the second slice by producing the line that goes through $f(1/n)$ and $f(2/n)$, but with the subtraction of the value produced by the previous unit. Thus in short:
$y_2(x)= n[f(1/n)-f(0)]x
-y_1(x)$. More generally, the output of the $k$-th linear unit 
 approximates the function $f$ in the $k$-th slice producing the line that goes through $f(k-1/n)$ and $f(k/n)$, but with the subtraction of $y_{k-1}$.
 [Note: as an alternative construction, the linear units could also be taken to be constant, with $y_k(x)=f(k-1/n)-y_{k-1}(x)$, and $y_1(x)=f(0)$.]

\end{proof}
 The same construction can be applied over any closed interval, as well over any finite union of closed intervals. Furthermore, if the range is $\R^p$, the same construction can be applied to each component. And finally, the same construction can be generalized if the input domain is of the form $[0,1]^m$. Thus in short:
 
\begin{theorem}
 Every continuous function $f$ from a compact set $C \subset \R^m$ to $\R^p$ can be approximated to any degree of precision $\epsilon$ by a shallow attention network comprising linear units gated by  corresponding linear threshold gate units, with a final dot product output. 
\end{theorem}

\subsection{Attention Layers: Dot Products}

\begin{figure}[ht]
\begin{center}
\includegraphics[width=1.0\columnwidth]{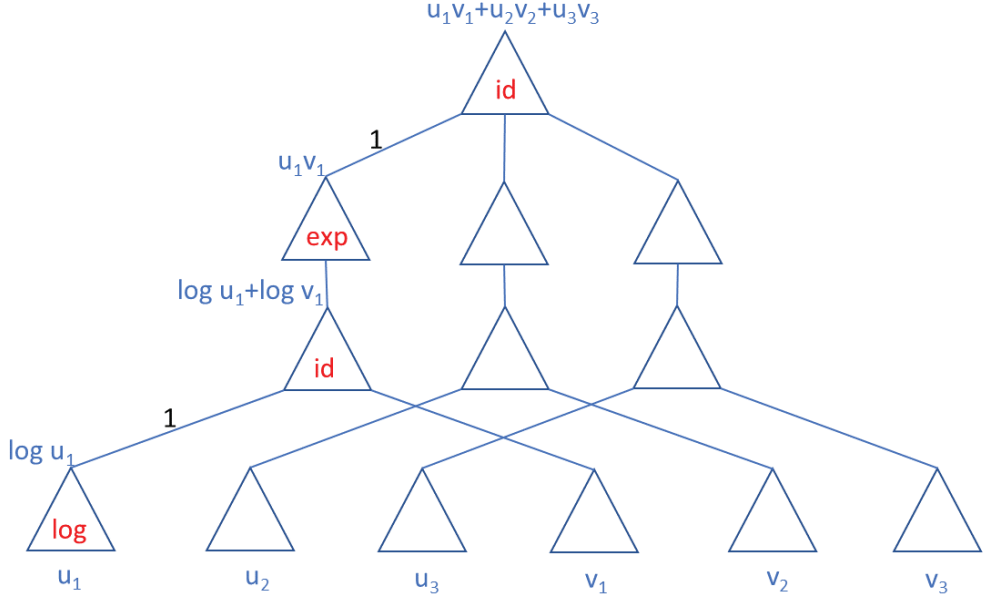}
\end{center}
\vspace{-0.1cm}
\caption{Standard model neural network for computing the dot product of two vectors $(u_1,u_2,u_3)$ and $(v_1,v_2,v_3)$ .}
\label{fig:DotProduct}
\end{figure}

\begin{figure}[ht]
\begin{center}
\includegraphics[width=1.0\columnwidth]{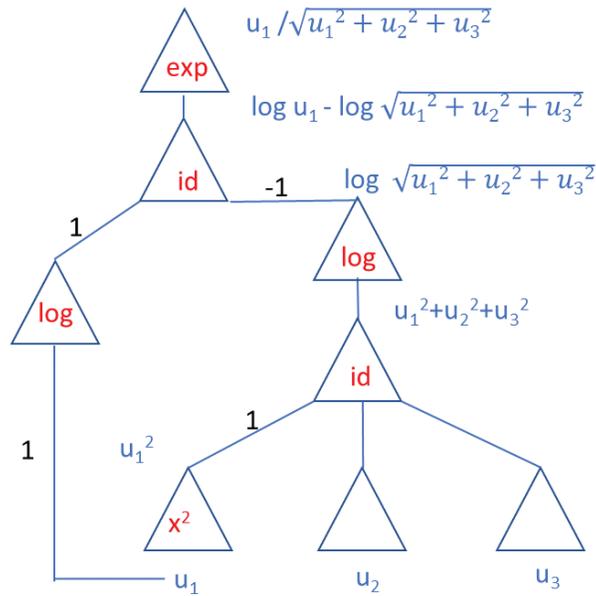}
\end{center}
\vspace{-0.1cm}
\caption{Standard model neural network for normalizing a vector $(u_1,u_2,u_3)$ (for clarity, only the first normalized component is fully shown).}
\label{fig:Normalization}
\end{figure}

\begin{figure}[ht]
\begin{center}
\includegraphics[width=1.0\columnwidth]{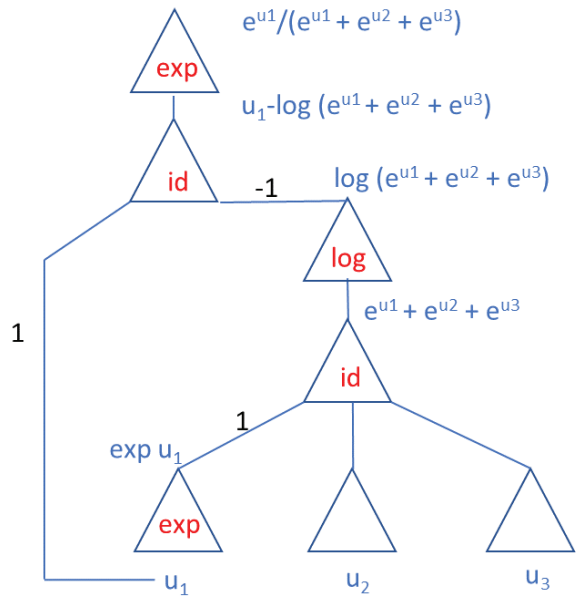}
\end{center}
\vspace{-0.1cm}
\caption{Standard model neural network for computing the softmax function for a vector $(u_1,u_2,u_3)$(for clarity, only the first component is fully shown).}
\label{fig:SoftMax}
\end{figure}

As we have seen in the section on transformers and the universal approximation proof above, one place where attention mechanisms are particularly important is for computing the dot product between two activity vectors $u=(u_1,\ldots,u_n)$ and $v=(v_1,\ldots, v_n)$, associated with two corresponding layers of $n$ neurons each. This can be achieved through output gating to first compute all the pairwise products $u_iv_i$ and then combine these products through a single linear output unit, with all its incoming weights set to $1$, to compute the dot product $uv= \sum_i u_iv_i$. However, this dot product can equally be computed by synaptic gating, i.e. by using the vector $v$ to gate the incoming weights of the linear unit above and compute the dot product in the form $uv= \sum_i (1.v_i) u_i$.
This can be scaled up to tensors where there are multiple output vectors $u(k)=(u^k_i)$
and multiple attention vectors $v(l)=(v^l_i)$ of the same length, and all pairwise dot products $u(k)v(l)$ are computed, for any $(k,l)$ pair, as in the transformer architectures. Of course the dot product can also be computed in the standard model (Figure 
\ref{fig:DotProduct}) however this requires a deeper network with four layers of standard units with fixed connections all equal to $1$, and both logarithm and exponential transfer functions. Thus
output or weight attention create a new primitive, or compact circuit, for computing dot products. The same is true of other operators that are often introduced in neural network without being part of the standard model, such as for the already-mentioned softmax 
(Figure \ref{fig:SoftMax}) or the
normalization of a vector
(Figure \ref{fig:Normalization}). 

\subsection{Attention Layers: Attention Weights}
Synaptic gating of a connection can suppress or enhance the corresponding incoming signal. 
Synaptic gating all the incoming edges of a unit allows to assign different importance
to its different inputs. In addition, it is often desirable that these degree of importance form a probability vector, as in the transformer architecture, and this can be achieved through a softmax operation.
It is possible to apply a normalizing softmax either to the vector of pairwise products $u_iv_i$, or to the rows or columns of the tensor of dot products $u(k)v(l)$, as in the transformer architectures.
The output of these softmax operations can then be used to gate other synaptic weights. These gated weights are often equal and  set to one in order to compute convex combinations, as in the transformer architecture
Thus, in short, in transformer and other architectures, attention mechanisms allow
dot products, softmax, and synaptic gating operations to be combined into one macro operation, which would require a network of depth $\sim 10$ for its implementation inside the SM.

\section{Cardinal Capacity Review}
\label{sec:cardinalreview}

We have seen that attention mechanisms enable important functionalities with minimal depth compared to the equivalent SM circuits, at the cost of adding attention neurons and mechanisms. Here we want to better understand the trade offs between the computations that are enabled and the corresponding costs. The key concept for doing so is the concept of cardinal capacity \cite{baldi2019capacity} which we briefly review below.

\subsection{Definition of Capacity:} Given a class of functions $\AA$, we define its cardinal capacity $C(\AA)$, or just capacity, to be: $C(\AA)= \log_2 \vert \AA \vert$, where $\vert \AA \vert$ is the cardinality of $\AA$ in the finite case.
In the continuous case, $\vert \AA \vert$ can be defined as a volume, but here we will focus primarily on finite cases. 
The class $\BB_n$ of all Boolean functions of $n$ variables has capacity $C(\BB_n)=2^n$.
Here we will consider subclasses of $\BB_n$, in particular those implemented by feed-forward networks of linear or polynomial threshold gates, with attention mechanisms, and compute the corresponding capacity.

\subsection{Linear and Polynomial Threshold 
Functions}
Linear or polynomial threshold functions are reasonably good approximation of linear- or polynomial-activation neurons with steep sigmoidal activation functions and, as such, are not particularly restrictive. 
A polynomial threshold functions of degree $d$ has the form $\sgn p(x)$,
where $p(x)$ is a polynomial of degree $d$ using a $-/+$ output representation. Alternatively, for a 0/1 output representation, we can use the form $H(p(x)$ where $H$ is the Heaviside function
equal to 0 for $x\leq 0$ and to 1 otherwise.
Units with values in $0/1$ are similar to logistic sigmoidal units, and units with values in $-1/+1$ are similar to 
$\tanh$ sigmoidal units.

We let $\TT(n;d)$ denote the class of polynomial threshold functions of degree $d$. Thus $\TT(n;1)$ denotes the class of linear threshold functions. 
When the inputs to a threshold function are binary, we use the term threshold gate.
In the case of polynomial threshold gates, it does not matter whether their input is encoded using $0/1$ or $-/+$ (or for that many any two distinct real numbers). This is because there is an affine transformation between any two such encodings and the affine transformation can be absorbed in the synaptic weights, i.e. the coefficients of $p$. The same is generally true for the encoding of the output, however when attention gating is considered the $0/1$ and $-/+$ encodings behave differently. For instance, in the case of output gating, the product of two $0/1$ threshold gates behaves like an AND, whereas the output of two $-/+$ gates behaves like an NXOR.

Thus to derive more general results, we will consider the case where the gating mechanism is implemented by a Boolean function $B$, which could be an AND, an NXOR, or something else. 
In the most general setting, we let $B(z_1,\ldots,z_k) : \{-1,1\}^k \to \{-1,1\}$ be a Boolean formula in $k$ variables. 
We are interested in the class of functions of the form 
$B(f_1,...,f_k): \{0,1\}^n \to \{-1,1\}$ where $f_j \in \TT(n;d_j)$. We denote this class by $\TT_B(n; d_1,\ldots,d_k)$.

\subsection{Why Capacity is Important}

The capacity $C(\AA)$ is a measure of what the class of functions $\AA$ can do. As a single number, it is of course a very crude representation of the true functional capacity. However in the case of neural networks the capacity has a stronger significance. To see this, note first that the cardinal capacity is also the number of bits required to specify an element of $\AA$. Thus in the case of neural networks, to a first order of approximation, the capacity is the number of bits that must be transferred from the training data to the synaptic weights during learning for the network to learn to implement a specific function in the class $\AA$.

\subsection{Capacity of Single Units: Review}

Before we estimate the capacity of single units with attention mechanisms, we must review the known capacity results on single units without attention mechanisms. 
For a single linear threshold gate of $n$ variables, we have \cite{zuev1989asymptotics,zuev1991combinatorial}:

\be
\left ( 1- \frac{10}{\log n} \right) n^2 \leq C(\TT(n;1)) \leq n^2
\label{eq:zuev}
\ee
This result was refined to the form \cite{kahn1995probability}:
\be
C(\TT(n;1))= n^2 - n \log_2n  \pm O(n)
\label{eq:komlos}
\ee
Similar results have been obtained for polynomial threshold gates of degree $d$
\cite{baldi88a,baldi2019polynomial}. In particular, for any $n$ and $d$ satisfying
$1 \leq d \leq n^\alpha$ (where $\alpha$ is fixed and $<\alpha <1$) there exists a constant $D=D(\alpha)$ such that: 

\be
(1-\frac{D}{\log n})^d n {n \choose \leq d}  \leq C(\TT(n;d))\approx n {n \choose \leq d}
\label{eq:poly100}
\ee
where:

\be
{n \choose \leq d} = \sum_{k=0}^d {n \choose k}
\ee
For degree $d=o (\log n)$, including fixed degree $d$, Equation \ref{eq:poly100} yields:

\be
C(\TT(n;d))= \frac{n^{d+1}}{d!}(1-o(1))
\label{eq:bv}
\ee

\subsection{Activation Attention and Muliplexing}
\label{sec:multiplexing}
We now describe one of the main techniques that will be used in the attention capacity proofs for both synaptic and output gating. Perhaps surprisingly, this technique can be viewed as a form of attention, specifically a form of activation attention or multiplexing. 
It was developed and used in 
\cite{baldi2018neuronal,baldi2019capacity}. First, we need the following lemma, stated for the 0/1 $n$-dimensional hypercube, but equally valid on the -/+ hypercube, or any other hypercube $[a,b]^n$. The lemma basically states that any vertex of the hypercube can be separated from the rest of the cube by a hyperplane with large margins.

\begin{lemma} \label{lm:multi}
Let $H$ be the $n$-dimensional hypercube, and $M>0$ and $K\geq 0$.
Fix any vertex $c=(c_1,\ldots, c_n)$ of the hypercube, and let 
$D=H-\{ c \} $. Then there exists affine linear functions of the form 
$f(x)=a_0+ \sum_{1}^n a_i x_i$ such that:
$f(c)=K$ and $f(d)\leq -M$ for any $d \in D$.
\end{lemma}

\begin{proof}
First note that there are 1:1 affine maps between the different hypercubes, thus it is enough to prove the result for the 0/1 hypercube. Second, all the corners play a symmetric role so it is enough to prove it for the corner $c=(1,1, \ldots,1)$.
It is easy to check that:
$f(x)=\sum_{1}^n (M+K)x_i -(M+K)n +K $ satisfies the conditions of the Lemma. Note that by using $-f(x)$ the sign of the regions and corresponding margins can be exchanged
($f(c)=-K$ and $f(d)>M$ for all $d \in D$). 
\end{proof}

\begin{figure}[ht]
\begin{center}
\includegraphics[trim={0 0 0 1.0cm},width=1.1\columnwidth]
{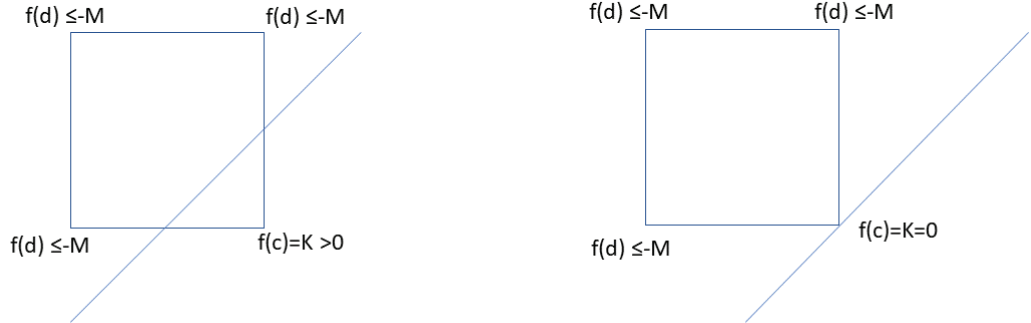}
\end{center}
\vspace{-1.3cm}
\caption{Any corner $c$ of the $n$-dimensional hypercube can be separated from all other corners $d$ of the hypercube by an affine hyperplanee with large margins defined by the parameters 
$K \geq 0$ and $M>0$.
}
\label{fig:Lemma}
\end{figure}

\begin{figure}[ht]
\begin{center}
\includegraphics[width=1.1\columnwidth]{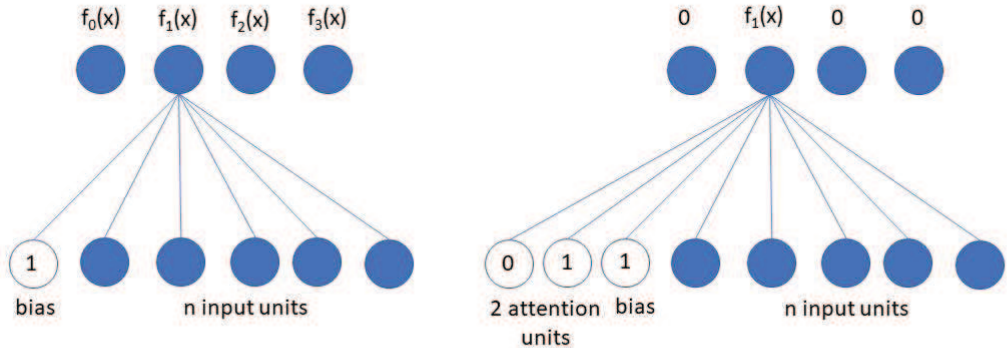}
\end{center}
\vspace{-1.3cm}
\caption{Left: A fully connected feedforward neural network with $n$ inputs and $m=4$ threshold gates computing the functions $f_0(x),f_1(x),f_2(x)$ and $f_3(x)$. The bias unit is constantly set to 1.
Right: The same network with $2=\lceil \log_2 m \rceil$ additional attention units in the input payer. The weights from the input units to the threshold gates are the same as in the left image. 
The attention units can be in 4 different states $(00)$, $(01)$, $(10)$, and $(11)$; these states can be associated in 1:1 fashion with the $m=4$ threshold units. Assume for instance that the state $(1,0) $ is associated with the hidden unit computing $f_1(x)$ (hidden unit 1). Then by Lemma \ref{lm:multi} applied to the hypercube of dimension 2, it is possible to choose a set of weights from the attention units and the bias to hidden unit 1 providing: (1) an attention activation of 0 when the attention units are in the (0,1) state; and (2) an arbitrarily large negative (or an arbitrarily large positive) attention activation for all the other 3 states.
As a result, when the attention units are in the (0,1) state the  output of the attended hidden unit 1 is equal to $f_1(x)$. When the attention units are in any of the other three states, unit 1 is not attended and its output is constant and equal to 0 (or constant and equal to 1). And similarly, mutatis mutandis, for the other three hidden units. In other words, we can first choose a fixed pattern of 0s and 1s in the hidden layer, called a mask, and then connect the attention units to the hidden layer with such weights that the output of the hidden layer is equal to the mask, except for one position associated with the attended unit. If the attended unit is unit $i$ in the hidden layer, the corresponding output is equal to $f_i(x)$ (and the attention units must be set to the corresponding values). 
}
\label{fig:Multiplexing}
\end{figure}

Now consider a neural network consisting of $n$  inputs fully connected to a hidden layer of $m$ linear or polynomial threshold functions (Figure \ref{fig:Multiplexing}) $f_0(x), \ldots , f_{m-1}(x)$. In the multiplexing approach, we add $m$, or even just $
\lceil \log_2 m \rceil $ new binary inputs to the input layer. $m$ different binary patterns over these inputs can be associated in one to one fashion with one of the $m$ threshold functions in the hidden layer. Let $i$ be any integer
$0 \leq i \leq m$ and let $p(i)$ denote the corresponding pattern of bits. For simplicity we can just use the binary representation of $i$, but any other representation works equally well. 

This pattern $p(i)$ can be viewed as a corner of the corresponding hypercube of dimension  $
\lceil \log_2 m \rceil$ and thus we can apply Lemma \ref{lm:multi} above to choose the weights connecting the attention units and the bias to hidden unit $i$ accordingly. In particular, the weights can be chosen such that:
(1) the attending signal originated from the attending bit patterns $p(i)$
is equal to 0; and (2) for all other settings of the attending bits, the attending signal is arbitrarily large and negative (alternatively arbitrarily large and positive). Ans similarly for all the other units and attention input patterns. 
As a result, whenever $p(i)$ appears in the attention bits, the $i$-th output of the hidden layer is equal to $f_i(x)$, and for all the other settings of the attention bits, the $i-th$ output is constantly equal to 1, or constantly equal to 0 (or -1 in the case of $-/+$ 
threshold hidden units). The pattern of constant bits is called the mask and different masks can be used for different proofs. Thus, in short, the attending signal emanating from the attention units is multiplexed with the regular signal and used to focus the attention of the hidden layer on the hidden unit encoded by the bits appearing in the attention units. The output of the hidden layer is equal to the mask except for the attended position, where it is equal to the corresponding function $f_i(x)$.

This form of activation attention is the key tool for proving capacity lower bounds. To see this, consider for instance the case where an OR operator is applied to the outputs of the hidden layer. With a mask consisting of 0s, when the attention bits are set to $p(i)$, the output of the OR applied to the hidden units is equal to $f_i(x)$. Thus the truth table of the overall input-output function of the original inputs plus the attention bits is uniquely equal to $f_i(x)$ when the attention bits are set to $p(i)$. Thus the capacity of the network with the expanded input of size $n+ \lceil log m \rceil$ is lower bounded by the sum of the capacities associated with the functions $f_i$ over the original input of size $n$.

\section{Capacity of Single Unit Attention}
\label{sec:capacitysingle}

\begin{figure}[ht]
\begin{center}
\includegraphics[trim={0 0 0 3.5cm},width=1.1\columnwidth]{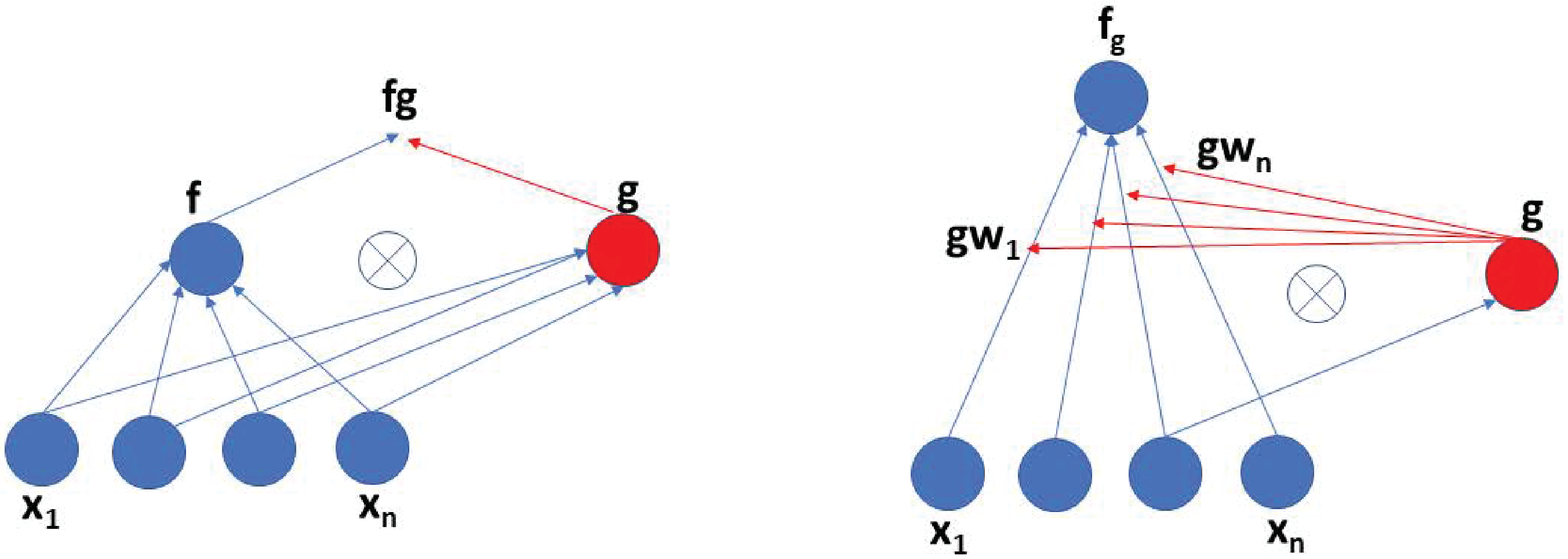}
\end{center}
\vspace{-1.3cm}
\caption{Left: Output gating with a single attention unit. Both the gated function $f$ and the gating function $g$ are linear (or polynomial) threshold gates of the same input vector $x=(x_1, \ldots, x_n)$. The overall network computes the function $fg(x)=f(x)g(x)$. Right: Synaptic gating with a single attention unit $g(x)$ gating all the incoming weights of the gated function $f(x)$. So the overall network computes the function $f_g(x)$. For instance,  if $f$ is a linear threshold gate $f(x)=\sign( \sum_i w_ix_i)$, then $f_g(x)=\sign (
\sum_i g(x) w_ix_i)$.
}
\label{fig:AttentionUnit12}
\end{figure}

We can now begin to estimate the capacity of various attention circuits, when the attention signal originate in a single gating unit, as shown in Figure \ref{fig:AttentionUnit12}.

\subsection{Capacity of Single Attention Units: Output Gating}

\begin{figure}[ht]
\begin{center}
\includegraphics[trim={0 0 0 1.0cm},width=0.99\columnwidth]{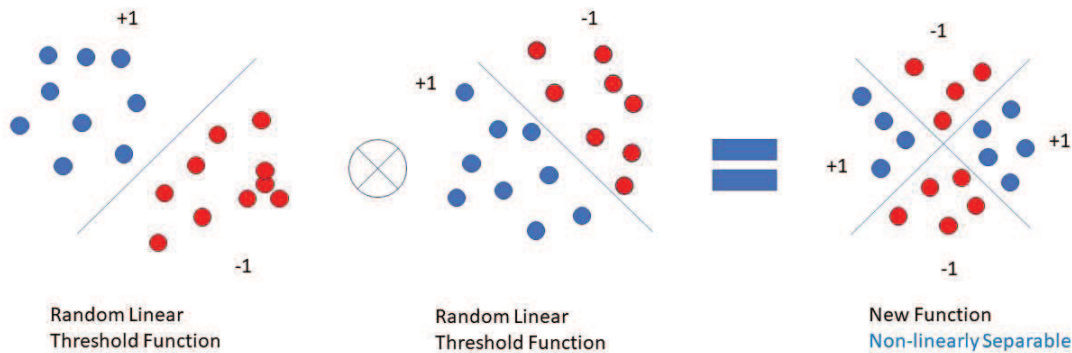}
\end{center}
\vspace{-1.3cm}
\caption{Two randomly selected -/+ linear threshold functions and their product.
We can randomly pick such functions by randomly picking normal vectors of weights $w=(w_i)$ on the unit sphere $S^{n-1}$ (or using i.i.d coordinates that are Normal or Uniform).
When $n$ is large, the normal vectors $w$ and $v$ are approximately orthogonal and the corresponding hyperplanes partition the space  into four regions, each one containing approximately $2^{n-2}$ points of the hypercube. In general, the resulting function is not linearly separable.}
\label{fig:capacity1}
\end{figure}

We want to compute the capacity of the 
class of all functions that can be computed by one neuron gated by another neuron, corresponding to the left hand side of Figure
\ref{fig:AttentionUnit12}. In the purely linear case,we have seen that this is the set of all quadratic functions of the form 
$O=(\sum_i w_ix_i)(\sum_j v_jx_l)$.
To partially address this question in the non-linear case, we can consider first the case of a linear threshold gate gated by another linear threshold gate, and then similarly for polynomial threshold gates of degree $d$. Using $-/+$ linear threshold gates for the gated and the gating units, this is the class of Boolean functions of the form:

\be
fg(x)=f(x)g(x) =\sign (\sum_i w_ix_i) \sign
(\sum_i v_i x_i)=\sign \left ( (\sum_i w_ix_i)(\sum_jv_jx_j) \right )
\label{eq:}
\ee
This class contains the identity and all the linear threshold gates. Thus, by Zuev's result (Equation \ref{eq:zuev}) its capacity is at least $n^2(1+o(1))$. However, intuitively, it must contain many other functions as shown in Figure \ref{fig:capacity1} suggesting that in general the product of two linearly separable functions is not linearly separable. On the other hand, the upperbound on the capacity is at most $2n^2(1+o(1)$, because the capacity is always bounded by the sum of the capacity of each individual component. Similarly considerations can be made for the $0/1$ which leads to the more general problem of estimating the capacity of the class of functions of the from $B(f,g)$ where $B$ is any Boolean operator, and $f$ and $g$ are linear or polynomial threshold gates.  
And even more generality can be obtained by considering
classes of Boolean functions of the form 
$B(f_1,\ldots, f_k)$ where $B$ is a $k$-ary Boolean operator and $f_1,\ldots,f_k$ are polynomial threshold gates of respective degrees $d_1,\ldots,d_k$. We first address the case of $k=2$ and then the general case.  

\subsubsection{Pairwise Composition ($k=2$).}
\label{seq:negative}

\begin{table}
  \caption{All possible Boolean combinations. There are 16 possible Boolean functions $B(p,q)$ of two variables $p$ and $q$. Each row corresponds to a function $B(p,q)$ and its negation $\lnot B(p,q)$. 
Ten Boolean functions are irreducible  i.e. they cannot be expressed as a function of a smaller number of variables. Eight Boolean functions are symmetric ($B(p,q)=B(q,p)$). Fourteen Boolean functions can be implemented by a linear threshold gate (LTG). The functions are organized into four groups separated by horizontal lines. Within a group, all the functions in the same column are equivalent when their arguments are implemented by linear threshold gates. The last column correspond to the cardinal capacity $C(B(f,g))$ when $f$ and $g$  vary among all possible linear threshold functions of the same $n$ variables.
 }
  \label{tab:boolean}
  \centering
  \begin{tabular}{llllll}
    \toprule
    $B(p,q)$    & $\lnot B(p,q)$     &Irred. ($k=2$) & Sym & LTG & C\{B(f,g)\}) \\
    \midrule
    $T$ & $F$   &no&  yes & yes & $1$     \\ \hline
    $p$ & $\lnot p$ & no& no & yes & $n^2(1+o(1))$      \\
    $q$ & $\lnot q$ & no &no & yes &    \\ \hline
    $p \, {\rm OR} \, q$ & $\lnot p \, {\rm AND} \, \lnot q$ &yes &  yes & yes &$2n^2(1+o(1))$ \\
    $p \, {\rm OR} \, \lnot q$ & $\lnot p \, {\rm AND} \,  q$ &yes&  no & yes & \\
      $\lnot p \, {\rm OR} \, q$ & $\ p \, {\rm AND} \, \lnot q$ &yes&  no & yes & \\
      $\lnot p \, {\rm OR} \, \lnot q$ & $ p \, {\rm AND} \,  q$ & yes & yes & yes & \\
      \hline
      $p\, {\rm XOR} \,q$ & $\lnot (p XOR q)$ &yes & yes & no & $2n^2 (1+o(1))$\\\hline
  \end{tabular}
\end{table}

For completeness, consider Table \ref{tab:boolean} summarizing all 16 Boolean function $B(p,q)$ of two variables. We can substitute $p$ and $q$ with arbitrary linear (or polynomial) threshold functions $f$ and $g$ and compute the corresponding cardinal capacity.
The first group in the table correspond to always true (T) and always false (F) functions, thus to a negligible total capacity of 1. The second group corresponds to a single linear threshold function, and thus its capacity is equal to:
$n^2 (1+o(1))$
All the elements in the second group in the table are also found in the third group corresponding to the AND and OR operators, because $f \, {\rm AND} \, f =f \, {\rm OR} \, f=f$.  Within the third group, all the OR expression are equivalent to each other, and all the AND expressions are equivalent to each other, when $p$ and $q$ are substituted with linear (or polynomial) threshold gates. This is because whenever $f$ is a polynomial threshold gate of degree $d$, then $\lnot f$ is also a polynomial threshold gate of degree $d$.
The first three groups cover 14 Boolean functions $B(p,q)$ in total. These 14 Boolean functions can be implemented by a single linear threshold gate of 
$p$ and $q$, and no other linear threshold gate of $p$ and $q$ exist. Thus the total aggregated capacity corresponding to all these cases, is given by the cardinal capacity $C(n,2,1)$ of a network of linear threshold gates with $n$ inputs, 2 hidden units, and 1 output unit. This capacity is given by \cite{baldi2018capacity,baldi2019capacity}:

\be
C(n,2,1) = 2n^2 (1+o(1))
\label{eq:}
\ee

There is a one-to-one correspondence between the set $ \{f \, AND \, g\}$ and the set $\{f \, OR \, g\}$ through the negation operator. Therefore:

\be
C(\{f \, {\rm AND} \, g\})=C(\{f \, {\rm OR} \, g\})  
\label{eq:}
\ee
[Note that any Boolean function that isolates once corner of the hypercube is irreducible. For such a function, knowing the values of the sequence $(B(f,g), B(f,\lnot g), B(\lnot f,g), B(\lnot f,\lnot g))$ uniquely  determines the values of $f$ and $g$].
The relevant result in \cite{baldi2018capacity,baldi2019capacity} is obtained using the attention multiplexing technique, applied in fact with the OR Boolean function and a mask of 0s, as described in Section
\ref{sec:multiplexing}. Thus, in short:

\be
C(\{f \, {\rm AND} \, g\})=C(\{f \, {\rm OR} \, g\})  =2n^2(1+o(1))
\label{eq:}
\ee

For the last row of the table, 
the output gating (multiplication) of two $-/+$ linear threshold functions correspond to applying the negation of the XOR Boolean operator. Note that:
$ f \, XOR \, \lnot g \equiv \lnot f \, XOR \, g \equiv
\lnot (f \, XOR \, g)$ and $f\, XOR \, g \equiv \lnot f\, XOR \, \lnot g$. As a result we have:

\be
\vert\{  f \, XOR \, g \} \vert = \vert \{\lnot (f\,  XOR \, g) \} \vert
\label{eq:}
\ee
when $f$ and $g$ vary over all possible linear threshold gates. Even more strongly, the corresponding sets of Boolean functions are identical:
\be
\{ f\, XOR \, g \}=\{\lnot (f \, XOR \, g) \}
\label{eq:}
\ee
Now it is easy to see that 
\be
n^2 (1 +o(1)) \leq C(\{ f \, XOR\, g \})=C(\{\lnot (f\, XOR \, g) \}) \leq 2n^2 (1+ o(1))
\ee
The lower bound is obtained by noticing that for any Boolean function $f$, $f \, XOR \, F=f$. The upperbound is obtained by noticing that 
$f \, XOR\, g$ can be implemented by 
a network $A(n,2,2,1)$ of linear threshold gates (using the disjunctive normal form), and the capacity of such a network is always at most equal to the sum of the capacities of its individual gates. Finally, the attention multiplexing technique described in Section
\ref{sec:multiplexing} applied with a mask of 1s (since $f \, NXOR \, T= f $)
shows that:

\be
C(\{ f XOR g \})=C(\{\lnot (f XOR g) \}) = 2n^2 (1+ o(1))
\ee
Thus the product of $0/1$ or $-/+$ linear threshold gates have the same capacity, and a similar argument holds for polynomial threshold gates. These results can be summarized in the following theorem, which is true for both $0/1$ and $-/+$ threshold gates:

\begin{theorem}
\label{thm:main10}
The capacity of a linear threshold gate output-gated by another linear threshold gate is given by: 
\be 
2n^2 \left (1+o(1) \right )
\label{eq:main101}
\ee 
Likewise,
the capacity of a polynomial threshold gate of degree $d$ output-gated  by another polynomial threshold gate of the same degree is given by:
\be 
2 \frac{n^{d+1}}{d!} \left ( 1+o(1) \right )
\ee
\end{theorem}

\begin{remark}
Furthermore, we have seen that every Boolean function can be written as a product of linear threshold gates with an exponential number of terms (Proposition \ref{prop:universal}). Theorem \ref{thm:main10} shows that it is not possible to do so using only a polynomial number of terms, since this would result in an overall capacity that is only polynomial, whereas the capacity of $B_n$ is $2^n$.
\end{remark}

\begin{remark}
The estimate in Equation \ref{eq:main101}
can be slightly refined using Equation 
\ref{eq:komlos} instead of \ref{eq:zuev}.
\end{remark}

\begin{remark}
These results can be extended to other interesting cases. For instance, if we assume that the weights of the gated and gating linear threshold neurons are binary with $-/+$ values,  then the output gating capacity is equal to 
$2n \left ( 1 + o(1) \right )$.
\end{remark}

\begin{figure}[ht]
\begin{center}
\includegraphics[width=0.99\columnwidth]{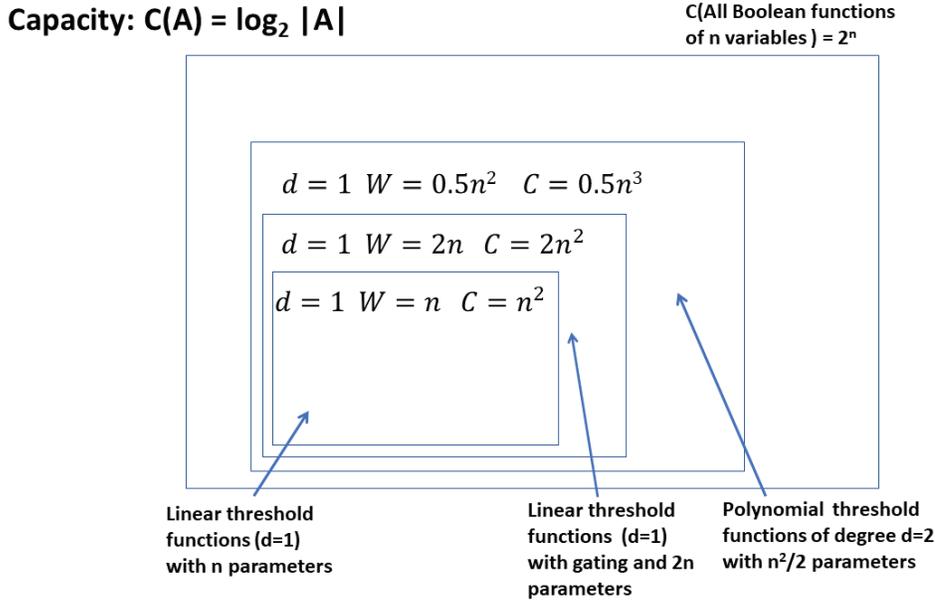}
\end{center}
\vspace{-0.2cm}
\caption{Hierarchy of classes of Boolean functions of $n$ variables according to their asymptotic capacity $C$ and number of parameters $W$. Linear threshold functions require $n$ parameters and achieve capacity $n^2$. Linear threshold functions gated by linear threshold functions require $2n$ parameters and achieve larger capacity (e.g. they contain XOR) equal to $2n^2$. Quadratic threshold functions require
$n^2/2$ parameters and achieve capacity  $n^3/2$
\cite{baldi2019polynomial}. The set of all Boolean functions correspond to an exponential capacity exactly equal to $2^n$. Note that in all these cases $C=nW$.}
\label{fig:hierarchy}
\end{figure}

\subsubsection{General Composition ($k\geq 2$).}
\label{seq:genral}

The results in the previous section can be generalized to the class of functions of the form $B(f_1,\ldots,f_k)$, where $B$ is a Boolean function of $k$ variables and, for each $j$, $f_j \in \TT(n;d_j)$. We denote this class by: $\TT_B(n; d_1,\ldots,d_k)$.

\begin{theorem}[Composition]	
\label{thm:composition}
  Let $B$ be an irreducible Boolean operator in $k$ variables.\footnote{Irreducibility means that $B$ can not be expressed as a Boolean operator in fewer than $k$ variables.}
  Then:
  \be
  \prod_{j=1}^k \abs{\TT(n-k+1;d_j)}
  \le \abs{\TT_B(n; d_1,\ldots,d_k)} 
  \le \prod_{j=1}^k \abs{\TT(n;d_j)}
  \ee
  Furthermore, if $B$ is the set of all irreducible Boolean functions of two variables 
  (there are 10 of them), we have:
      \be
  \abs[2]{\bigcap_B \TT_B(n; d_0,d_1)} 
  \ge \abs{\TT(n-1;d_0)} \, \abs{\TT(n-1;d_1)}
  \label{eq:}
  \ee
  where the intersection is over the ten irreducible binary Boolean operators.
    \label{eq:th1}
  \end{theorem}
  
The complete proof of this theorem is given in the Appendix. The upper bound is easy and the lower bound relies on the attention multiplexing approach (Section \ref{sec:multiplexing}). To check the special case, when $k=2$, for two polynomial threshold gates of degree $d_1$ and $d_2$, Theorem \ref{thm:composition} yields:

\be
  \abs{{\TT(n-1;d_1)}{\TT(n-1;d_2)}}
  \le \abs{\TT_B(n; d_1,d_2)} 
  \le \abs{{\TT(n;d_1)}{\TT(n;d_2)}}
  \ee
and when $d_1=d_2=d$:
\be
  \abs{\TT(n-1;d)}^2
  \le \abs{\TT_B(n; d,d)} 
  \le \abs{\TT(n;d)}^2
  \ee
Thus, in the case of output gating of two linear threshold gates, we have:
\be
  \abs{\TT(n-1;1)}^2
  \le \abs{\TT_B(n; 1,1)} 
  \le \abs{\TT(n;1)}^2
  \ee
Substituting the estimates in Equations \ref{eq:zuev}--\ref{eq:bv} in these inequalities gives immediately Theorem
\ref{thm:main10}. Note that the intersection across all 10 irreducible Boolean functions is large.

\subsection{Capacity of Single Attention Units: Synaptic Gating}

We are now ready to compute the capacity for the case corresponding to the right hand side of Figure \ref{fig:AttentionUnit12}, where one threhsold unit synaptically gates the weights of another threshold unit. 
To begin with, we look at the case where all the weights of the gated unit are gated simultaneously. The main result is as follows:

\begin{theorem}
\label{thm:fullsynapticgating}
Let $f(x)$ and $g(x)$ be two linear or polynomial threshold gates (not necessarily of the same degree), both with the same $-/+$ or $0/1$ output encoding and $n$ binary input variables. Then full synaptic gating of $f$ by $g$, where all the coefficients of $f$ are multiplied by $g$, is equivalent to output gating of $f$ by $g$. In particular, if both gates are linear threshold gates, then the corresponding capacity is given by:
\be
2n^2 \left ( 1 +o(1) \right )
\ee
and if both gates are polynomial threshold gates of degree $d$, then the corresponding capacity is given by:
\be
2\frac {n^{d+1}}{d!} \left ( 1 +o(1) \right )
\ee
\end{theorem}

\begin{proof}
We sketch the proof when $f$ and $g$ are linear threshold gates, but the argument extends immediately to polynomial threshold gates. Let us assume that $f(x) = \sign (\sum_i w_ix_i)$ and $g(x) = \sign (\sum v_ix_i)$. Then, with full synaptic gating, the gated function satisfies:
$f_g(x)=\sign (\sum_i g(x)w_i x_i)=
\sign(g(x) \sum_i (w_i x_i)=\sign g(x)
\sign (\sum_i w_i x_i) = g(x) f(x)$.
In the case of $0/1$ units, if $H$ is the Heaviside function, then: 
$f_g(x)=H (\sum_i g(x)w_i x_i)=H(g(x)
\sum_i w_i x_i)$. If $ g(x)=1$, this is the same as $f(x)g(x)$. Likewise if $g(x)=0$, as long as we define $H(0)=0$, then $f_g(x)$ is also equal to: $f(x)g(x)$. [Note that the gating is applied to the bias too].
\end{proof}

\begin{remark} In this particular case, to some extent, we can also consider the mixed case. If $f(x)$ is a $-/+$ gate and $g(x)$ is a $0/1$ gate, if we define $\sign 0 =0$ then we also have $f_g(x)=f(x)g(x)$ everywhere. If $f(x)$ is a $0/1$ gate and $g(x)$ is a $-/+$ gate, then when $g(x)=1$ we also have $f_g=fg$. However, when $g(x)=-1$, then $f_g(x)=\lnot f(x)=1-f(x)$.
\end{remark}

\begin{figure}[ht]
\begin{center}
\includegraphics[width=0.70\columnwidth]{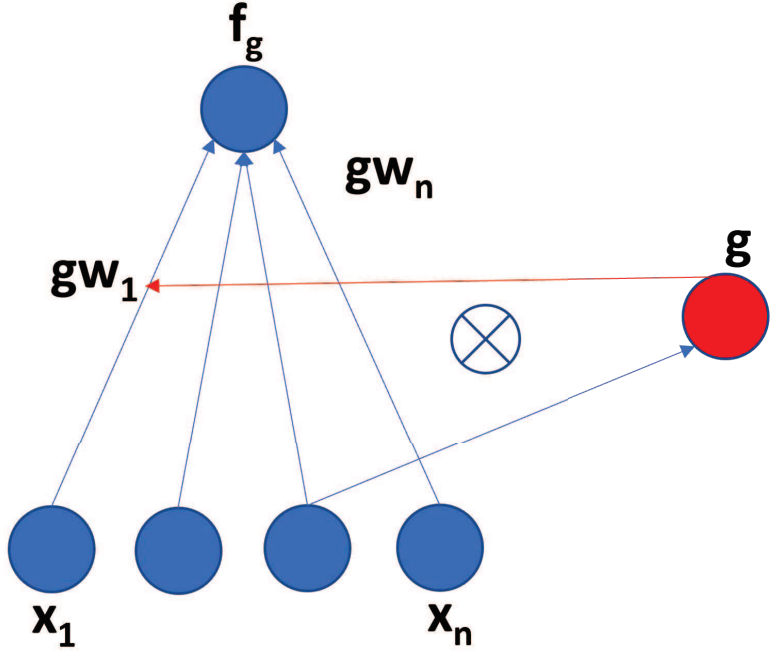}
\end{center}
\vspace{-0.2cm}
\caption{Synaptic gating with a single attention unit. Both the gated function $f$ and the gating function $g$ are linear (or polynomial) threshold gates of the same input vector $x=(x_1, \ldots, x_n)$.  Through the synaptic gating operation, a single synaptic weight of the function $f$ ($w_1$ in the figure) is multiplied by $g$.}
\label{fig:synapticgatingoneweight}
\end{figure}

\begin{remark}
In both Theorems \ref{thm:main10} and \ref{thm:fullsynapticgating}
there is approximately a doubling of the capacity at the cost of doubling the number of parameters.
\end{remark}

Finally, we consider the synaptic gating case  where the gating unit gates only {\it one} of the weights of the gated unit
(Figure \ref{fig:synapticgatingoneweight}). 
The following Proposition provides bounds on the corresponding capacity. 

\begin{proposition}
\label{prop:singleweightgating}
Consider the case of a linear threshold gate $f$ with $n$ binary inputs, where one of the weights is synaptically gated by the output of a second linear threshold gate $g$ of the same $n$ inputs.
Then the capacity $C$ satisfies:
\be
n^2 \left ( 1 + o(1) \right ) \leq  C \leq 2n^2\left ( 1 + o(1) \right )
\ee
If the linear threshold gates are replaced by polynomial threshold gates of degree $d$, the capacity $C$ satisfies:
\be
\frac{n^{d+1}}{d!} \left ( 1 + o(1) \right ) \leq  C \leq 2       
\frac{n^{d+1}}{d!}
\left( 1 + o(1) \right )
\ee
The same bounds hold for the case of additive activation attention between two linear polynomial threshold gates, or two linear polynomial threshold gates, of the same $n$ inputs. 
\end{proposition}

\begin{proof}
The result is true for both $0/1$ and $-/+$ encodings of the outputs. We provide the proof in the linear case but the technique is the same for polynomial threshold gates of degree $d>1$. The lower bound results immediately from the fact that the gating unit could have an output constant and equal to 1 $g(x)=1$). In this case the gated function is equal to $f(x)$ and the lower bound is the corresponding capacity estimate. 
The upperbound is simply the sum of the capacities. A similar argument applies for the case of additive activation attention. 
\end{proof}

\section{Capacity of Attention Layers}\label{sec:capacitylayers}
\begin{figure}[ht]
\begin{center}
\includegraphics[trim={0 0 0 1.0cm},width=0.99\columnwidth]{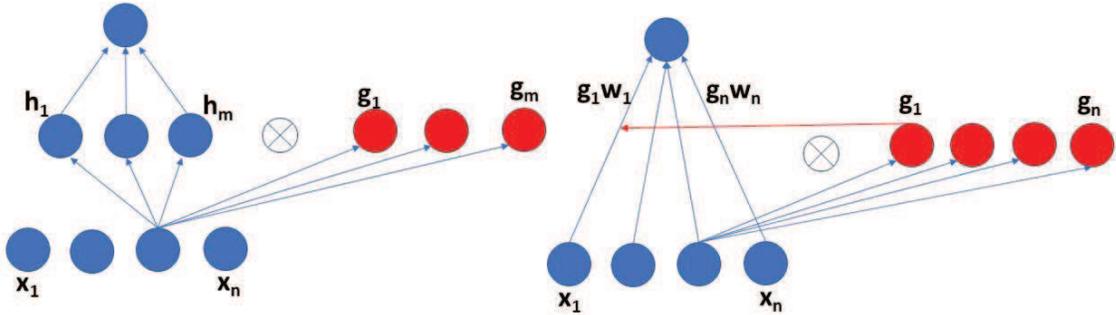}
\end{center}
\vspace{-1.0cm}
\caption{Left: output gating by a gating layer. For the same $n$ dimensional input 
vector $x$, there are $m$ hidden units computing functions $h_1(x),\ldots, h_m(x)$,
and $m$ corresponding gating units computing functions
$g_1(x),\ldots, g_m(x)$. With the gating, the effective output of the hidden units is given by $h_1(x)g_1(x), \ldots, h_m(x)g_m(x)$. The final output unit produces an output of the form $f(h_1(x)g_1(x), \ldots, h_m(x)g_m(x))$.
In the capacity analysis, we assume that the functions $h$, $g$, and $f$ are linear threshold gates. 
Right: synaptic gating by a gating layer.
In this case, there is a unit computing 
a function $f(x)$ with $n$ weights $w_i, \ldots, w_n$. There are $n$ gating functions 
$g_1(x),\ldots, g_n(x)$, each one multiplicatively gating one of the weights $w$. If $f= \sign (\sum_i w_ix_i)$
then $f_g(x)=\sign (\sum _i g_i(x) w_i x_i)$.
}
\label{fig:AttentionLayer12}
\end{figure}

The previous attention results are obtained using only two neurons, a gating neuron and a gated neuron, with either output gating or synaptic gating. We now extend the capacity analysis to  cases where there is a layer of gating neurons, as shown in Figure \ref{fig:AttentionLayer12} for both output and synaptic gating.

\subsection{Capacity of Attention Layers: Output Gating}

We now examine the capacity of a network with one attention layer with output gating, as depicted on the left hand side of Figure \ref{fig:AttentionLayer12}. Thus we consider an architecture with $n$ inputs, $m$ hidden linear threshold units gated by $m$ corresponding linear threshold units, and one final linear threshold output gate. All the linear threshold gates have $-/+$ outputs, although the following theorem is unchanged, and the method of proof is similar, if the gates have 0/1 outputs. We denote by
$\TT({n,m,1};\times)$ the corresponding set of Boolean functions. 
Note that this is the same architecture for computing the dot product of the gated and the gating hidden layer outputs, except that the final unit is non-linear with variable weights, instead of being linear with fixed weights equal to one.
We will also let $\TT({n,1};\times)$ denote the set of Boolean functions corresponding to one linear threshold gate of $n$ variables output-gated by another linear threshold gate of the same variables.  

\begin{theorem}
\label{thm:outputlayergating}
The capacity $C(\TT({n,m,1};\times))$ of the
set of Boolean functions corresponding to $n$ inputs, $m$ hidden linear threshold gates output-gated by $m$ hidden linear threshold gates of the same inputs, followed by one linear threshold gate output satisfies:

\be
mn^2    \leq C(\TT({n,m,1};\times)) \leq   2mn^2 \left ( 1+o(1) \right )
\label{eq:}
\ee
for $n \to \infty$,
and for any choice of $m \in [1,2^{o(n)}]$.
Furthermore:

\be 
C(\TT({n,m,1};\times))=m C(T(n,1;\times)) \left (1+o(1) \right)
\ee
Thus: 
\be 
 C(\TT({n,m,1};\times)) =   2mn^2 \left (1+o(1) \right )
 \ee
\end{theorem}

\begin{proof}
Let us denote by $f$ the map between the input layer and the hidden layer with gating, and by $\phi$ the map from the hidden layer to the output layer. For the upper bound, we first note that the total number of possible maps $f$ is bounded by $2^{mC(\TT(n,1;\times))}\leq
2^{2mn^2(1+o(1))}$, since $f$ consists of $m$ threshold gates gated by $m$ threshold gates, and thus each gated unit corresponds to at most $2^{C(\TT(n,1;\times))}\leq  2^{2n^2(1+o(1))}$ possibilities by the Theorems in Section \ref{sec:capacitysingle}.
Any fixed map $f$, produces at most $2^n$ distinct vectors in the hidden layer. It is known \cite{anthony2001discrete} that the number of threshold functions $\phi$ of $m$ variables defined on at most $2^n$ points is bounded by:
\be
2 {2^{n}-1 \choose \leq m} =2^{nm (1+o(1))}
\label{eq:}
\ee
using the assumption $m \leq 2^{o(n)}$. Thus, under our assumptions, the total number of functions of the form $\phi \circ f$ is bounded by the product of the bounds above which yields immediately:
\be
C(\TT({n,m,1};\times) )\leq
mC(\TT(n,1;\times)) \left (1+o(1) \right )
\leq 2mn^2 \left (1+o(1) \right )
\label{eq:}
\ee
For the lower bound, we can force the gating units to be the identity (i.e. with a constant output equal to 1). In this particular case, the gating units can be ignored and we need to count the number of Boolean functions that can be implemented in the remaining architecture. A theorem in \cite{baldi2019capacity} shows that this number is equal to $mn^2(1+o(1))$.

To prove the rest of the theorem, we use 
attention multiplexing. As a reminder, the basic idea is to have a small set of the input units act as attention units that can be used to select a particular function in the hidden layer. The same setting of the attention units will be used to select the corresponding functions in both the gating and gated layers. 
More formally, we decompose $n$ as: $n=n^- + n^+$  where $n^- = \lceil \log_2 m \rceil$ corresponds to the attention units. Likewise, we decompose each input vector  $x=(x_1,\ldots,x_n )\in \{-1,+1\}^n$ as: $x=(x^-,x^+)$, where:
\be
 x^-=(x_1,\ldots,x_{n^-}) \in
\{-1,+1\}^{n^-} \quad {\rm and} \quad 
 x^+=(x_{n^-+1}1,\ldots,x_{n}) \in
\{-1,+1\}^{n^+} 
\label{eq:}
\ee
For any gated Boolean linear threshold map $f^+$ from
$\{ -1,+1\}^{n^+}$ to $\{-1,+1\}^m$, we can uniquely derive a map
$f=(f_1,\ldots, f_m)$ from $\{ -1,+1\}^{n}$ to $\{-1,+1\}^m$ defined by:
\be
f_i(x^-,x^+)=  [x^-=i] \;\; AND \;\; [f_i^+(x^+)]
\label{eq:}
\ee
Here $x^-=i$ signifies that the binary vector $x^-$ represents the digit $i$. In other words $x^-=i$ is used to select the $i$-th unit in the gated layer as well as in the gating layer, 
and filter $f^+$ by retaining only the value of $f_i^+$. By Lemma \ref{lm:multi}),
this selection procedure can be expressed using a single linear threshold function of the input $x^-$ for the gated layer, and similarly for the gating layer. 
We say that $f$ is obtained from $f^+$ by multiplexing and $f$ is a gated threshold map.
It is easy to see that the filtering of two distinct maps $f^+$ and $g^+$ results into two distinct maps $f$ and $g$.
Now let us use $\phi = OR$ in the top layer--note that OR can be expressed as a linear threshold function. Then it is also easy to see that $\phi \circ f \not = \phi \circ g$. Thus the total number of Boolean functions that can be implemented in this architecture is lower-bounded by the number of all gated Boolean maps $f^+$. This yields:

\be
C(\TT(n,m,1;\times)) \geq m
C(\TT(n^+,1;\times)) \left (1 + o(1) \right )=
2m n^2 \left (1 + o(1) \right )
\label{eq:}
\ee
using the fact that $n^+=n- \lceil \log_2 m \rceil$,
and $\lceil \log_2 m \rceil = o(n)$ by assumption.
Thus:
$C(\TT(n,m,1;\times))=mC(\TT(n,1;\times)) \left ( 1 + o(1) \right)=2mn^2 \left ( 1+o(1) \right )$. 
\end{proof}

\begin{remark}
In Theorem \ref{thm:outputlayergating}, we see again that both the capacity and the number of parameters approximately double at the same time.  
\end{remark}

\subsection{Capacity of Attention Layers: Synaptic Gating}

We now examine the capacity of a network with one attention layer with synaptic gating, as depicted on the right hand side of Figure \ref{fig:AttentionLayer12}, with each gating neuron gating a different weight of a gated neuron.

\begin{proposition}
Consider the case of a linear threshold gate with $n$ inputs and $n$ weights, where each weight is synaptically gated by an independent linear threshold gate of the same $n$ inputs.
Then the capacity $C$ satisfies:
\be
n^2 \left ( 1 + o(1) \right ) \leq  C \leq n^3\left ( 1 + o(1) \right )
\ee
If the linear threshold gates are replaced by polynomial threshold gates of degree $d$, the capacity $C$ satisfies:
\be
\frac{n^{d+1}}{d!} \left ( 1 + o(1) \right ) \leq  C \leq        
\frac{n^{d+2}}{d!}
\left( 1 + o(1) \right )
\ee
\end{proposition}

\begin{proof}
The proof is similar to the proof of Proposition \ref{prop:singleweightgating}. The lower bound is obtained by constraining all the gating units to have a constant output equal to 1. The upperbound is simply the sum of all the capacities. 
\end{proof}

Likewise, we can consider an architecture with $n$ inputs, one layer of $m$ gating units, and one parallel layer of $m$ gated units. Each gating unit is uniquely paired with one gated unit (one to one) and synaptically gates one of the weights of the gated unit.

\begin{proposition}
Consider the case of 
an architecture with $n$ inputs, one layer of $m$ gating units, and one parallel layer of $m$ gated units. Each gating unit is uniquely paired with one gated unit (one to one) and synaptically gates one of the weights of the gated unit. 
Then the capacity $C$ satisfies:
\be
m n^2 \left ( 1 + o(1) \right ) \leq  C \leq 2m n^2\left ( 1 + o(1) \right )
\ee
If the linear threshold gates are replaced by polynomial threshold gates of degree $d$, the capacity $C$ satisfies:
\be
m \frac{n^{d+1}}{d!} \left ( 1 + o(1) \right ) \leq  C \leq        
2m\frac{n^{d+1}}{d!}
\left( 1 + o(1) \right )
\ee
\end{proposition}

\begin{proof}
The proof is similar to the proof of Proposition \ref{prop:singleweightgating}. The lower bound is obtained by constraining all the gating units to have a constant output equal to 1. The upperbound is simply the sum of all the capacities. 
\end{proof}

\section{Conclusion}
\label{sec:conclusion}


In addition to the fundamental role attention plays in brain function, attention mechanisms have also become important for artificial neural networks and deep learning. 
Here we have taken the first steps towards building a theory of attention mechanisms by first identifying the quarks of attention, i.e. its smallest building blocks.
Using the three variable types of the SM allows for the systematic identification and organization of possible attention building blocks based on their origin type, target type, and whether the mechanism of action is additive or multiplicative. Assuming that the attention signal originates from the output of some neurons, this yields six possibilities, which can then be reduced to three main cases: activation attention, output gating, and synaptic gating. Activation attention falls within the SM, whereas output gating and synaptic gating correspond to multiplicative extensions of the SM. Current attention-based architectures in deep learning, including transformers, are built out of attention modules which are themselves built out of output gating and synaptic gating operations. These operations and modules can be viewed as new primitives in the language of neural architectures in digital simulations and, because they are differentiable, the usual backpropagation learning framework can easily be extended to them.
However, in a physical neural machine, these operations require additional connections (wires) and physical mechanisms for implementing multiplicative interactions. 

Ouput gating can be used dynamically to directly silence unattended neurons, and to magnify the output of attended neurons. It can also be used as the main building block of a shallow module that can compute the dot product of two vectors of neuronal activities. The latter is a key, massively used, component of transformer architectures. 

Synaptic gating is a fast synaptic mechanism that can be used dynamically to silence or weigh the attended synapses. It is often used in combination with a softmax operator to enable dynamic convex combinations of vectors, as in the transformer architectures. 
The concept of fast synapses that can vary their strengths on fast time scales is not new and has been associated with different roles,  
in different contexts. For instance, one potential role is the storage of transient information, such as intermediary results during mental reasoning, or simply the memorization of the beginning of a paragraph as the reading of the paragraph proceeds. A second potential role stems from viewing synaptic weights as computer programs, and thus fast synapses as enabling dynamic changes in the programs that are being executed and the implementation of parameterized functions. And a third role studied here is the enabling of attention. These three roles are not independent and raise interesting architectural
questions for deep learning and neuroscience and the possible need for multiple synaptic time scales interacting in hierarchical ways.

To see this, as an example, consider the reading paradigm where information about the first sentence of a long paragraph is stored using a set of fast weights. If, as the reading proceeds, one must suddenly access 
a specific subset of this transiently stored information, 
attention must be directed towards certain particular words contained in the first sentence. 
In a deep learning architectures, this can be thought of in terms of a softmax synaptic gating, as is done in transformer and other NLP architectures. Thus somehow this fast weight attention mechanism must operate upon, and be faster than, the fast weight synaptic mechanism used to store information about the first sentence.

Attention mechanisms allow the attending network to modulate the function computed by the attended network, thereby expanding the scope of useful functions that can be efficiently implemented and trained in deep learning. 
Because the SM already has universal approximation properties, its extensions should not be evaluated in terms of which functions can be approximated, but rather in terms of other efficiencies. 
While attention blocks act as new primitives in standard deep learning software libraries, 
having access to output gating and synaptic gating mechanisms in a physical neural network can reduce its depth. 
Using the notion of cardinal capacity, and working with the approximation provided by Boolean neurons (linear or polynomial threshold gates), enables systematic investigations of the capacity of attentional circuits that were previously not possible. In particular, we have been able to estimate the capacity of basic attentional circuits involving linear, or polynomial, threshold gates.
In many cases of interest, we found essentially a doubling of the capacity with a doubling of the number of parameters, which is a sign of efficiency. 

Perhaps surprisingly, a key ingredient in the capacity proofs is the third form of attention, activation attention. Activation attention is used to prove capacity lower bounds by the multiplexing approach which selects a unit in a layer, as a function of the attending units, while driving the remaining units in the layer to low or high saturation.  
There is work left for tightening some of the estimates and for extending them to other activation functions and other architectures.

\begin{figure}[ht]
\begin{center}
\includegraphics[trim={0 0 0 0.5cm},width=0.99\columnwidth]{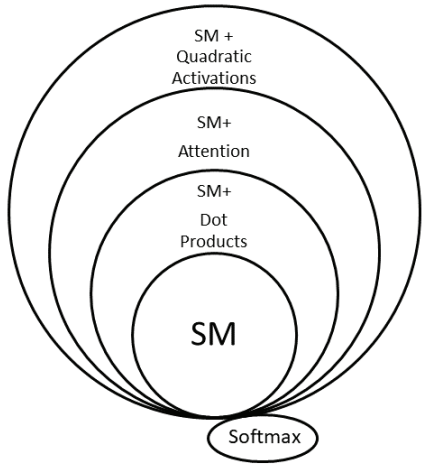}
\end{center}
\vspace{0cm}
\caption{Standard model and some of its extensions.}
\label{fig:SM}
\end{figure}

Overall, both output and synaptic gating are extensions of the SM which introduce quadratic terms 
in the SM (Figure \ref{fig:SM}). Quadratic terms are powerful but expensive: a neuron with full quadratic activation over its $n$ inputs requires on the order of $n^2$ synaptic parameters. Using quadratic activations everywhere in a large deep architecture leads to implementations that
may not be efficient in terms of parameters and learning. Attention mechanisms are a way of introducing quadratic terms in a sparse way, in order to gain some of the benefits of quadratic activations, without paying the full price.

Finally, we can return to the quotes in the introduction linking attention to awareness and pointing to the inadequacy of having a single term.
While subjectively we feel that we can control and direct our attention and be aware of its shifts, it should be obvious that attention mechanisms, such as output or synaptic gating, are computational mechanisms that do not require awareness. They can operate at all levels of a cognitive architecture, for instance to help implement dynamically whole-part hierarchies and ultimately awareness itself. Thus, in short, awareness is not necessary for attention, but attention may be necessary for awareness. Having a single term is indeed inadequate and, in time, it may have to be replaced with multiple terms
to better capture the underlying complexities.

\section{Appendix: Detailed Proof of Theorem \ref{thm:composition}}

Here a polynomial threshold function is a function of the form 
$f = \sign(p): \{0,1\}^n \to \{-1,1\}$ 
where $p$ is a polynomial in $n$ real variables of degree at most $d$.
The class of all such functions is denoted $\TT(n;d)$.

Let $B(z_1,\ldots,z_k) : \{-1,1\}^k \to \{-1,1\}$ be a Boolean function in $k$ variables. 
We are interested in the class of functions of the form 
$B(f_1,...,f_k): \{0,1\}^n \to \{-1,1\}$ where $f_j \in \TT(n;d_j)$. Denote this class by $\TT_B(n; d_1,\ldots,d_k)$.
We want to prove the following theorem:
\par\null\par
\noindent
{\bf Theorem} (Composition). {\it	
  Let $B$ be an irreducible Boolean operator in $k$ variables.
  Then:}
  \be
  \prod_{j=1}^k \abs{\TT(n-k+1;d_j)}
  \le \abs{\TT_B(n; d_1,\ldots,d_k)} 
  \le \prod_{j=1}^k \abs{\TT(n;d_j)}
  \label{eq:th1a}
  \ee

The upper bound is trivial from considering the total number of tuples $(f_1,...,f_k)$ with $f_j \in \TT(n;d_j)$. The lower bound is nontrivial except for $k=1$ where both bounds become identical.
The key to the proof is the multiplexing (activation attention) procedure, where $k$ input components are viewed as attention units capable of producing a constant mask in the hidden layer, except for the attended function. Here for simplicity we use a sparse encoding in the $k$ components, although dense encoding is also possible, as in the proof of Theorem \ref{thm:outputlayergating}. Dense encoding would lead to a reduction in the number of attending units from $k$ to $ \lceil \log_2k \rceil $ as in Section 
\ref{sec:multiplexing}. As a side note, using more attention units than the minimal number required, can be used to reduce the size of the attention weights, or to make the attention mechanism less sensitive to each individual attention bit.

To prove the lower bound in Composition Theorem~\ref{thm:composition}, let us restate it equivalently as: 
\be
\prod_{j=0}^k \abs{\TT(n-k;d_j)}
\le \abs{\TT_B(n; d_0,\ldots,d_k)} 
\le \prod_{j=0}^k \abs{\TT(n;d_j)}.
  \label{eq:th1.0}
\ee

\par\noindent
Irreducibility implies that if we select any input component $i$, the value of $B$ cannot be determined entirely from the value of the remaining components alone.  More formally:
 
\begin{lemma}		\label{lem: restriction}
  Consider an irreducible Boolean operator $B = B(z_0,\ldots,z_k)$ 
  and an index $i \in \{0,\ldots,k\}$.
  There exist signs $\theta \in \{-1,1\}$ and $\theta_j \in \{-1,1\}$, $j \in \{0,\ldots,k\} \setminus \{i\}$, such that:
  \be
  B(z_0,\ldots,z_k) = \theta z_i
  \quad \text{whenever } z_j = \theta_j
  \text{ for all } j \ne i.
  \label{eq:lm1}
  \ee
\end{lemma}

\begin{proof}
Consider $B(z_0,\ldots,z_k)$ as a function of $z_i$. If this function is constant in the variable $z_i$ no matter how we fix the other variables, then the value of $B(z_0,\ldots,z_k)$ is entirely determined by the values of these other variables, which contradicts irreducibility. Therefore, there exists some assignment $z_j = \theta_j$, $j \ne i$, so that
the function $B(\theta_0,\theta_1,\ldots, z_i, \ldots \theta_k)$
is not constant in $z_i$.  But there exists only two non-constant Boolean functions $f(x)$ in one variable: $f(x)=x$ or $f(x)=-x$, and this determines $\theta$. 
\end{proof}

\par\noindent
The next lemma essentially states that we can fit an affine function of $k$ variables to $k+1$ points.            
\par\noindent
\begin{lemma}		\label{lem: fitting}
  Let $e_0=0$ and $e_1,\ldots,e_k$ denote the canonical basis vectors in $\R^k$.
  Then, for any choice of index $j \in \{0,\ldots,k\}$ 
  and signs $\theta_i \in \{-1,1\}$, $i \in \{0,\ldots,k\} \setminus \{j\}$  
  there exists an affine function $q: \R^k \to \R$ such that:
  \be
  q(e_i) = 
  \begin{cases}
    0, & i=j \\
    \theta_i, & i \ne j
  \end{cases}
  \label{eq:lm2}
  \ee
 for all $i \in \{0,\ldots,k\}$.
\end{lemma}

\begin{proof}
It is straightforward to check that the function: 
\be
q(z) = \theta_0 - \theta_0 z_j + \sum_{i \in \{0,\ldots,k\} \setminus \{j\}} (\theta_i-\theta_0)z_i
\label{eq:}
\ee
satisfies the required property.
\end{proof}

\par\noindent
We can now use the previous lemma to derive a lemma for consistently extending a function of $n-k$ variables to a function of $n$ variables. Here $k$ components are used as selector of filter variables, as in the proof of Theorem 
\ref{thm:outputlayergating}.

\begin{lemma}			\label{lem: F}
  Consider a function $f \in \TT(n-k;d)$, an index $j \in \{0,\ldots,k\}$, 
  and signs $\theta \in \{-1,1\}$ and $\theta_i \in \{-1,1\}$, $i \in \{0,\ldots,k\} \setminus \{j\}$.  
  There exists a function $F \in \TT(n;d)$ such that:
 \be
  F(e_i \oplus x) = 
  \begin{cases}
    \theta f(x), & i=j \\
    \theta_i, & i \ne j
  \end{cases}
  \label{eq:}
  \ee
  for all $x \in \{0,1\}^{n-k}$. Here $\oplus$ denotes the concatenation operator.
\end{lemma}

\begin{proof}
Express the polynomial threshold function $f$ as: 
\be
f(x) = \sign(p(x))
\quad \text{for } x \in \{0,1\}^{n-k}
\label{eq:}
\ee
where $p$ is a polynomial in $n$ variables and of degree at most $d$.
Let $q$ be a function that satisfies the conclusion of Lemma~\ref{lem: fitting}.
Fix a number $M$ large enough so that $M > \abs{p(x)}$ for all $x \in \{0,1\}^{n-k}$,
and define:
\be
F(z \oplus x) = \sign \left( M q(z) + \theta p(x) \right)
\label{eq:}
\ee
for all $z \in \R^k$ and $x \in \R^{n-k}$.
By construction, $F$ is a polynomial threshold function on $\{0,1\}^n$ of degree at most $d$ as required.

Let us check that $F$ satisfies the conclusion of the lemma. 
If $z=e_j$, we have $q(z)=0$ due to our choice of $q$ (per the conclusion of Lemma~\ref{lem: fitting}), and we get 
$F(z \oplus x) = \sign(\theta p(x)) = \theta f(x)$.
If $z=e_i$ with $i \ne j$, then our choice of $q$ implies
$F(z \oplus x) = \sign(M \theta_i + \theta p(x))$.
The choice of $M$ guarantees that the term $M \theta_i$ dominates the term $\theta p(x)$ in magnitude, so we have
$F(s \oplus x) = \sign(M \theta_i) = \theta_i$.
\end{proof}

\par\noindent
We can now use Lemma \ref{lem: F} for the simultaneous extension and filtering of several functions of $n-k$ variables relative to an irreducible Boolean function $B$.

\begin{lemma}	\label{lem: embedding}
  For any $(k+1)$-tuple of functions $(f_0,\ldots,f_k)$ where $f_j \in \TT(n-k;d_j)$
  there exists a $(k+1)$-tuple of functions $(F_0,\ldots,F_k)$ where $F_j \in \TT(n;d_j)$
  such that:
  \be
  B(F_0,\ldots,F_k)(e_i \oplus x)
  = f_i(x)
  \label{eq:}
  \ee
  for all $i \in \{0,\ldots,k\}$ and $x \in \{0,1\}^{n-k}$.
\end{lemma}

\begin{proof}
Lemma~\ref{lem: restriction} yields the existence of signs $\theta_i \in \{-1,1\}$ for $i \in \{0,\ldots,k\}$ 
and $\theta_{ij} \in \{-1,1\}$ for distinct $i,j \in \{0,\ldots,k\}$, such that:
\begin{equation}	\label{eq: B restricted}
B(z_0,\ldots,z_k) = \theta_i z_i
  \quad \text{whenever } z_j = \theta_{ij}
  \text{ for all } j \ne i.
\end{equation}
Now consider the functions $f_j \in \TT(n-k;d_j)$, $j \in \{0,\ldots,k\}$.
Lemma~\ref{lem: F} yields the existence of functions $F_j \in \TT(n;d_j)$, $j \in \{0,\ldots,k\}$, such that:
\begin{equation}	\label{eq: Fj constructed}
F_j(e_i \oplus x) = 
  \begin{cases}
    \theta_i f_i(x), & i=j \\
    \theta_{ij}, & i \ne j
  \end{cases}
\end{equation}
for all $i,j \in \{0,\ldots,k\}$ and $x \in \{0,1\}^{n-k}$.

For any fixed $i \in \{0,\ldots,k\}$ and $x \in \{0,1\}^{n-k}$, by construction the variables $z_j \coloneqq F_j(e_i \oplus x)$ satisfy the condition in \eqref{eq: B restricted}. 
Therefore, \eqref{eq: B restricted} and \eqref{eq: Fj constructed} yield:
\be
B(F_0,\ldots,F_k)(e_i \oplus x) = B(z_0,\ldots,z_k) = \theta_i z_i 
= \theta_i F_i(e_i \oplus x)
= \theta_i^2 f_i(x)
= f_i(x)
\label{eq:}
\ee
as claimed.
\end{proof}

\medskip
\par\noindent
Armed with this lemma, we can now prove Theorem
\ref{thm:composition}.

\begin{proof}[Proof of Theorem~\ref{thm:composition}]
Lemma~\ref{lem: embedding} demonstrates that
for any tuple of functions $(f_0,\ldots,f_k) \in \prod_{i=0}^k \TT(n-k;d_j)$
there exists a function $F \in \TT_B(n;d_0,\ldots,d_k)$
such that $F(e_i \oplus x) = f_i(x)$
for all $i \in \{0,\ldots,k\}$ and $x \in \{0,1\}^{n-k}$.
Thus, each component $f_i$ of the original $k$-tuple can be uniquely recovered 
from $F$. Therefore, a map $(f_0,\ldots,f_k) \mapsto F$ (if there are multiple $F$ corresponding to some $f$, select one arbitrarily) defines an injection from the cartesian product $\prod_{i=0}^k \TT(n-k;d_j)$ into $\TT_B(n;d_0,\ldots,d_k)$, completing the proof. 
\end{proof}


As shown in Table \ref{tab:boolean},
there are $16$ binary Boolean operators $B$. Ten of them are irreducible, including AND, OR and XOR and their negations. For each such operator, the Composition Theorem 
\ref{thm:composition} gives:

\be
\abs{\TT(n-1;d_0)} \, \abs{\TT(n-1;d_1)}
\le \abs{\TT_B(n; d_0,d_1)} 
\le \abs{\TT(n;d_0)} \, \abs{\TT(n;d_1)}
\label{eq:}
\ee
Surprisingly, the intersection of all ten classes is still as large. 

\begin{proposition}
  We have: 
  \be
  \abs[2]{\bigcap_B \TT_B(n; d_0,d_1)} 
  \ge \abs{\TT(n-1;d_0)} \, \abs{\TT(n-1;d_1)}
  \label{eq:}
  \ee
  where the intersection is over the ten irreducible binary Boolean operators.
\end{proposition}

In particular, there are many functions $f$ (specifically, $2^{2n^2(1-o(1))})$
that can be simultaneously expressed
as: $f = f_1 \, \AND \, f_2 = f_3 \, \OR \, f_4=f_5 \, XOR \,  f_6$ where all the $f_i$ are linear threshold gates.

\begin{proof}
In the proof of the Composition Theorem~\ref{thm:composition} above,
we showed that for each irreducible Boolean operator $B$ 
and pair of functions $(f_0,f_1) \in \TT(n-1;d_0) \times \TT(n-1;d_1)$, 
there exists $F \in \TT_B(n; d_0,d_1)$ such that:

\be
F(0 \oplus x) = f_0(x), \quad 
F(1 \oplus x) = f_1(x)
\label{eq:}
\ee
for all $x \in \{0,1\}^{n-1}$.
Obviously, this pair of equations defines $F$ uniquely on $\{0,1\}$, and
$F$ is independent of $B$. Thus, $F$ lies in the intersection of $\TT_B(n; d_0,d_1)$
over all irreducible $B$.
\end{proof}

\section*{Acknowledgment}

Work in part supported by 
ARO  grant 76649-CS and NSF grant 1633631 to PB, and 
AFOSR grant FA9550-18-1-0031 to RV.

\bibliography{baldi,nn,Alex}
\bibliographystyle{plain}

\end{document}